\documentclass[10pt,twocolumn,letterpaper]{article}

\usepackage{cvpr}              

\usepackage[accsupp]{axessibility} 
\definecolor{cvprblue}{rgb}{0.21,0.49,0.74}
\usepackage[pagebackref,breaklinks,colorlinks,allcolors=cvprblue]{hyperref}
\usepackage{multirow}
\usepackage{makecell}
\usepackage{fontawesome}
\usepackage{xcolor}         
\usepackage{colortbl}
\usepackage{amsmath}
\def\paperID{} 
\def\confName{CVPR}
\def\confYear{2026}

\title{AdaptVision: Efficient Vision-Language Models via Adaptive Visual Acquisition}

\author{
Zichuan Lin$^{*\dagger}$ \quad 
Yicheng Liu$^*$ \quad 
Yang Yang \quad 
Lvfang Tao \quad 
Deheng Ye$^{\ddagger}$ \\
Tencent Hunyuan \\
{\tt\small lzcthu12@gmail.com, yichengliu.e@outlook.com, ydyl1991@gmail.com} \\
Code and models: \href{https://github.com/AdaptVision/AdaptVision}{github.com/adaptvision/adaptvision}
}

\begin{document}
\maketitle
\renewcommand*{\thefootnote}{\fnsymbol{footnote}}
\footnotetext{$^*$Equal contribution. $^\dagger$Project lead. $^\ddagger$Corresponding author.}
\begin{abstract}
Vision-Language Models (VLMs) have achieved remarkable success in visual question answering tasks, but their reliance on large numbers of visual tokens introduces significant computational overhead. While existing efficient VLM approaches reduce visual tokens through fixed-ratio compression, they operate passively and lack the ability to adapt to varying task requirements. This motivates a fundamental question: Can VLMs autonomously determine the minimum number of visual tokens required for each sample? Inspired by human active vision mechanisms, we introduce AdaptVision, an efficient VLM paradigm that enables adaptive visual token acquisition through a coarse-to-fine approach. Our model initially processes compressed visual tokens from low-resolution images and selectively acquires additional visual information by invoking a bounding box tool to crop key regions when necessary. We train AdaptVision using a reinforcement learning framework that carefully balances accuracy and efficiency. Central to our approach is Decoupled Turn Policy Optimization (DTPO), which decouples the learning objective into two components: (1) tool learning, which optimizes correct tool utilization, and (2) accuracy improvement, which refines the generated responses to improve answer correctness. Based on this formulation, we further decouple advantage estimation by computing separate advantages for tokens associated with each objective. This formulation enables more effective optimization for AdaptVision compared to vanilla GRPO. Comprehensive experiments across multiple VQA benchmarks demonstrate that AdaptVision achieves superior performance while consuming substantially fewer visual tokens than state-of-the-art efficient VLM methods. 
\end{abstract}    
\section{Introduction}

\begin{figure}[t]
    \begin{center}
        \includegraphics[width=1\columnwidth]{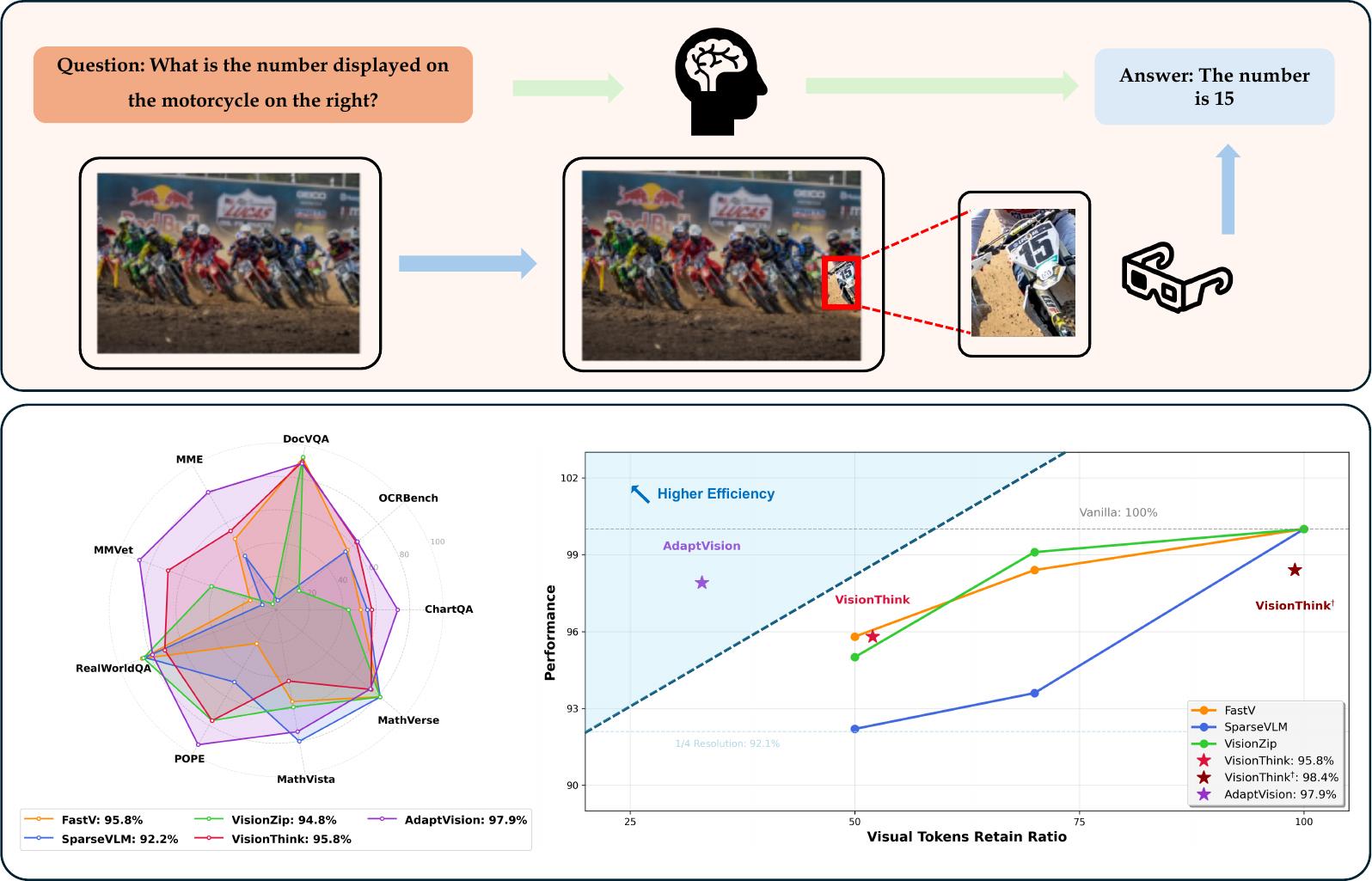}
    \end{center}
    \caption{\textbf{Our key motivations and AdaptVision performance and efficiency.} \textbf{Top: Coarse-to-fine.} Human visual attention mechanisms first guide the search for question-relevant regions in images, which are then subjected to detailed analysis. \textbf{Down:} AdaptVision achieves superior performance with significantly fewer visual tokens than previous efficient VLM methods.}
    \label{fig:intro}
\end{figure}

Recently, Vision-Language Models (VLMs)~\citep{li2023blip,bai2023qwen,chen2024sharegpt4v} have achieved significant breakthroughs in general visual question answering (VQA) and diverse practical applications by projecting and adapting visual tokens into large language model (LLM) space~\citep{touvron2023llama,achiam2023gpt,zhu2023minigpt,bai2023qwen}. However, the promising performance of VLMs largely relies on the large amount of vision tokens, inevitably introducing a huge memory and computational overhead when compared to LLMs, particularly for high-resolution images. For instance, a $2048 \times 1024$ image yields 2,678 vision tokens in Qwen2.5-VL~\citep{bai2025qwen2}. Therefore, it is crucial to avoid the excessive consumption of visual tokens.

Numerous studies have explored visual token compression to enhance VLM efficiency~\citep{yang2025visionzip,chen2024image,wen2024efficient,shi2023upop,he2024zipvl,jian2023expedited,zhang2024sparsevlm,yang2025visionthink}. Existing works can be categorized into two main research directions. 
The first prunes or merges a fixed number of visual tokens based on predetermined thresholds, according to the importance and similarity of vision tokens~\citep{yang2025visionzip,chen2024image,zhang2024sparsevlm}. 
The second dynamically processes distinct samples, where the system adaptively switches between using 100\% vision tokens for OCR-related tasks and 25\% vision tokens for simpler tasks by selectively employing quarter-resolution images~\citep{yang2025visionthink}. 
However, existing efficient VLM paradigms and methods are largely \textit{passive}, as they can only reduce the number of vision tokens by predefined ratios. 
This leads to a natural question: \textit{Can VLMs adaptively determine the minimum number of vision tokens for each sample according to different scenarios?}
Cognitive neuroscience reveals that our visual system operates 
through an active, sequential, and adaptive process known as \textit{active vision}~\citep{findlay2003active,itti2005neurobiology}. 
It first captures coarse, low-spatial-frequency information to grasp the gist of a scene, then directs attention to salient regions for detailed analysis~\citep{navon1977forest}.
This \textit{coarse-to-fine} processing mechanism enables humans to efficiently parse complex visual inputs with minimal cognitive load. Fig.~\ref{fig:intro} provides an illustrative example.


The cognitive strategy of active vision is operationalized in recent VLMs through the ``thinking-with-images'' paradigm, such as invoking tools to zoom and crop specific regions~\citep{zheng2025deepeyes, lai2025mini} to advance fine-grained visual understanding. We argue that this active reasoning capability can be effectively applied to the critical task of visual token reduction, allowing the model to decide how few visual tokens are sufficient.


In this paper, we propose AdaptVision, a framework that leverages visual tool use to determine the minimum visual token usage while maintaining high accuracy.
Our model initially processes compressed visual tokens from low-resolution images and adaptively acquires additional visual tokens by invoking a bounding box tool to crop key regions from the original high-resolution image when necessary. 
The model is trained via reinforcement learning~\citep{schulman2017proximal,lin2019unified,jiang2025multi,lin2023sample,lin2018episodic,yang2019fully} to balance accuracy and efficiency. 


However, training this dual-objective policy with standard RL algorithms like Group Relative Policy Optimization (GRPO)~\citep{shao2024deepseekmath} presents two key challenges: 
(1) \textit{Ambiguous credit assignment}: Vanilla GRPO assigns a single sequence-level reward to all generated tokens, failing to distinguish the contribution of the decision to request additional visual tokens from that of generating the final answer;
(2) \textit{Imbalanced optimization}: Since vanilla GRPO normalizes all tokens uniformly in a sequence, it introduces an imbalance: compared to 1-turn direct-answer sequences, 2-turn tool-invoking sequences receive imbalanced gradient signals, causing the latter to be under-optimized.

To address these challenges, we propose Decoupled Turn Policy Optimization (DTPO). 
First, to mitigate optimization imbalance, we decouple the learning objective into two components based on the functional roles of response tokens: (1) \textit{tool learning}, which encourages correct tool use, and (2) \textit{accuracy improvement}, which refines the generated responses to improve answer correctness. 
Each objective is normalized separately to balance learning signals across different tokens.
Second, to enable precise credit assignment, we decouple advantage estimation by computing distinct advantages for tokens associated with each objective, encouraging more efficient tool exploration.
Experiments on multiple VQA benchmarks demonstrate that AdaptVision achieves superior performance with significantly fewer visual tokens than state-of-the-art efficient VLM methods, as shown in Fig.~\ref{fig:intro}. 





In summary, our contributions are:
\begin{enumerate}[leftmargin=*]
    \item We introduce AdaptVision, a VLM framework that leverages visual tool use for dynamic token reduction.
    \item We propose a Decoupled Turn Policy Optimization (DTPO) algorithm alongside a tailored reward function to enable the effective training of AdaptVision.
    \item Extensive evaluation on multiple VQA benchmarks shows that AdaptVision achieves superior performance with substantially reduced visual token consumption compared to existing efficient VLM methods.
\end{enumerate}

\begin{figure*}[t]
\vspace{-0.5cm}
    \begin{center}
        \includegraphics[width=\textwidth]{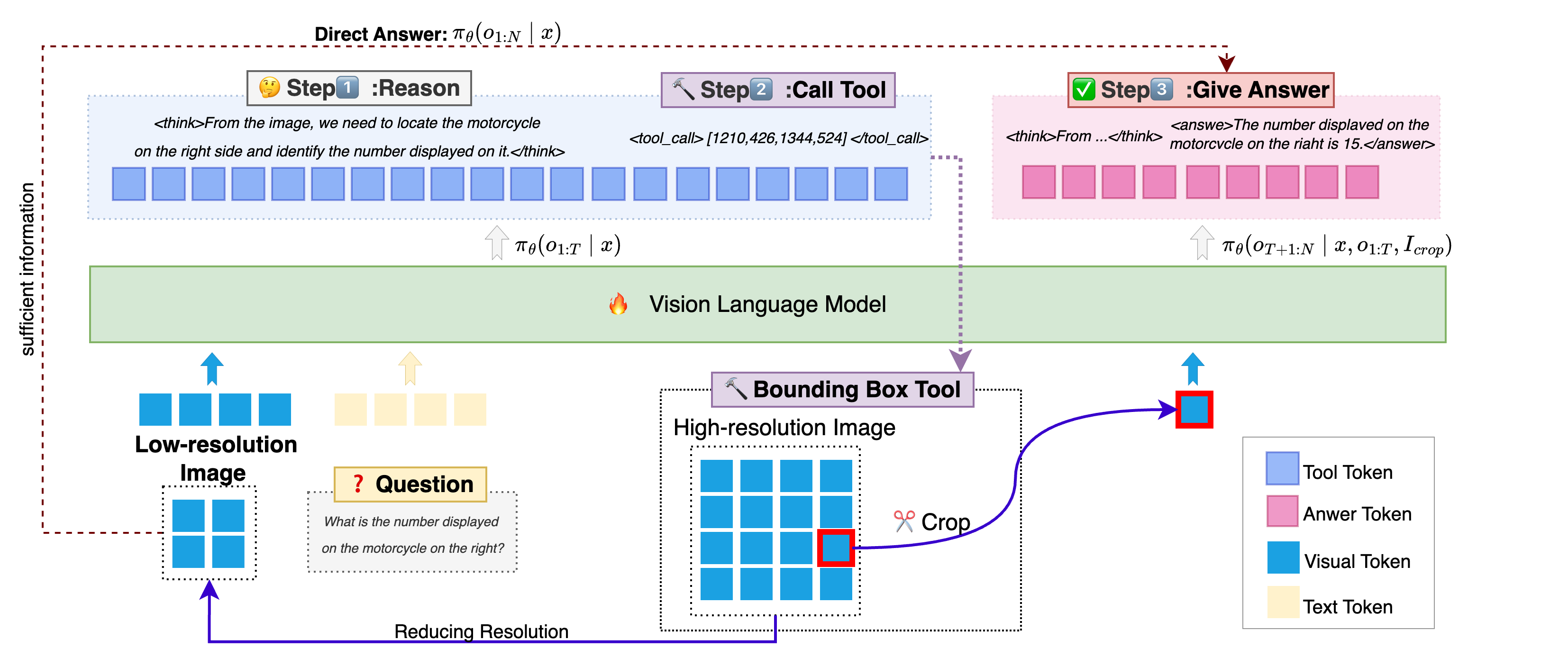}
    \end{center}
    \caption{\textbf{FrameWork of AdaptVision.} AdaptVision first processes a 1/4-resolution image. The model then decides whether to answer directly or invoke the bounding box tool to crop a high-resolution region for further analysis before generating the final answer.
    }
    \label{fig:framework}
\end{figure*}



\section{Related work}

\paragraph{Vision Language Model with Reasoning.}


Recent advances in reasoning LLMs such as OpenAI's o1~\citep{jaech2024openai} and DeepSeek R1~\citep{guo2025deepseek} have accelerated the use of RL to enhance reasoning capabilities~\citep{yang2025entropic}. This trend has extended to VLMs~\citep{tan2025reason,shen2025vlm,liu2025visual,peng2025lmm,wang2025seenav}, where most work focuses on high-level semantic reasoning like tool use or chain-of-thought explanation. A related direction explores active perception, equipping VLMs with fine-grained control mechanisms~\citep{wu2024v, huang2025high, su2025pixel}. Recent systems such as DeepEyes~\citep{zheng2025deepeyes} and Mini-o3~\citep{lai2025mini} support operations like zoom and crop, improving performance on detailed visual tasks. 
While these methods showcase the power of active visual reasoning for enhancing answer accuracy, how to apply this ``thinking with images" paradigm to the goal of computational efficiency, specifically for visual token compression, remains less explored. Our method enables the VLM to autonomously determine the minimum number of visual tokens required for a given task, thereby achieving efficient inference while maintaining performance.

\paragraph{Efficient VLM with Vision Token Compression.} 

Reducing VLM computational cost by vision token compression has become a popular research topic~\citep{kong2025token}. Existing methods rely on predefined rules or metrics to compress tokens. For instance, FastV~\citep{chen2024image} prunes a fixed 50\% of tokens based on attention scores after the second layer. PyramidDrop~\citep{xing2024pyramiddrop} proposes progressive token compression to reduce information loss. Other works leverage cross-modal relevance for token selection, such as SparseVLM~\citep{zhang2024sparsevlm} and VisionZip~\citep{yang2025visionzip}, which retain semantically relevant visual tokens. A key limitation of these methods is their dependence on a fixed compression ratio, which lacks adaptability across tasks. VisionThink~\citep{yang2025visionthink} uses RL to decide whether to use a low-resolution or the original image, offering limited adaptability but still restricting the model to coarse-grained decisions. In contrast, our approach enables the VLM to learn coarse-to-fine ability and adaptively determine the minimum number of visual tokens for each task.

\section{Preliminary}

\subsection{Reinforcement Learning for LLMs}

Recent studies~\citep{guo2025deepseek,jaech2024openai} have demonstrated RL effectively enhances the reasoning capabilities of large language models (LLMs). Recently, Group Relative Policy Optimization (GRPO)~\citep{shao2024deepseekmath} has been widely used in LLM reasoning. 
Given a question $x$, GRPO generates $G$ distinct responses $\{o_i\}_{i=1}^G$ with sequence length $N_i$ from the current policy $\pi_{\theta_{old}}$ and obtains a group of rewards $\{R_i\}_{i=1}^G$. 
GRPO optimizes the policy model $\pi_\theta$ by maximizing the following objective:
\begin{small}
\begin{multline}
    \mathcal{J}_{\text{GRPO}}(\theta) =  \mathbb{E}_{x, o_i}  \\
    \left[
        \frac{1}{G} \sum_{i=1}^G 
        \left(
            \frac{1}{N_i} \sum_{t=1}^{N_i} \mathcal{L}_{i,t}(\theta) 
             -  \beta \mathbb{D}_{\text{KL}} \left[ \pi_\theta(\cdot | x) \,||\, \pi_{\text{ref}}(\cdot | x) \right] 
        \right)
    \right]
\end{multline}
\end{small}
where $\mathcal{L}_{i,t}(\theta)$ denotes the token-level loss given by:
\begin{small}
\begin{multline}\label{eq:grpo}
    \mathcal{L}_{i,t}(\theta) = \min 
    \Bigg( 
        \frac{\pi_\theta(o_{i,t} \mid x,o_{i,<t})}{\pi_{\theta_{\text{old}}}(o_{i,t} \mid x,o_{i,<t})} A_{i,t},  \\
        \text{clip}  \bigg( 
        \frac{\pi_\theta(o_{i,t} \mid x,o_{i,<t})}{\pi_{\theta_{\text{old}}}(o_{i,t} \mid x,o_{i,<t})}, 1 - \epsilon, 1 + \epsilon \bigg) A_{i,t}
    \Bigg),
\end{multline}
\end{small}
\begin{equation}
A_{i,t} = \frac{R_i - \text{mean}(\{R_i\}^G_{i=1})}{\text{std}(\{R_i\}^G_{i=1})}, 
\end{equation}
\begin{equation}
\mathbb{D}_{\text{KL}} (\pi_\theta || \pi_{\text{ref}}) = \frac{\pi_{\text{ref}}(o_i|q)}{\pi_\theta(o_i|q)} - \log \frac{\pi_{\text{ref}}(o_i|q)}{\pi_\theta(o_i|q)} - 1,
\end{equation}
where 
$\mathbb{D}_{\text{KL}}$ is the KL-divergence measure. $\epsilon$ and $\beta$ are hyperparameters. The advantage estimate $A_{i,t}$ is computed using a group of rewards $\{R_i\}_{i=1}^G$.



\subsection{Vision Language Models}

The VLM architectures generally consist of three components: a visual encoder, a modality projector, and a LLM. A commonly used approach for the visual encoder is to employ a pre-trained image encoder like CLIP-VIT~\citep{radford2021learning} that converts input images into visual tokens. The modality projector adjusts the size of these visual tokens to match the embedding size of LLM and to achieve semantic alignment, enabling the LLM to process visual data effectively. The LLM then integrates the aligned visual and textual information to generate responses.

Existing works have revealed that the computational complexity of VLM is strongly influenced by the sequence length~\citep{yang2025visionzip}, where the sequence length is defined as $n = n_{sys} + n_{img} + n_{question}$. In typical VLM tasks, the number of vision tokens $n_{img}$ is often much larger than the other two, sometimes by a factor of 20. Therefore, reducing the number of vision tokens is the key for improving the efficiency of VLMs.



\section{Methodology}

\subsection{Framework}



We aim to develop an efficient VLM that minimizes visual token usage while maintaining high performance by adaptively acquiring visual information based on question and image complexity. As shown in Fig.~\ref{fig:framework}, our method first processes a low-resolution image ($I_{low}$), cutting visual token usage to 25\% of the original. The VLM then autonomously decides whether to answer directly or crop key regions ($I_{crop}$) from the high-resolution image for more detail.
As shown in Fig.~\ref{fig:framework}, given a low-resolution image $I_{low}$ and the question $q$, the model can output a \textit{direct answer} or invoke a \textit{tool call} using \texttt{<tool\_call>[$x_1, y_1, x_2, y_2$]</tool\_call>} to obtain $I_{crop}$ before reasoning further and answering.

However, the VLM lacks a mechanism for deciding which response style is most appropriate for a given input $x = \{ x_{sys}, I_{low}, q \}$. We therefore frame this as a reinforcement learning problem to optimize the following policy:
\begin{equation} \label{eq:pi}
\begin{aligned}
    & \pi_{\theta}(o | x) = \\
    & \begin{cases}
        \pi_{\theta}(o_{ 1: N} \mid x), & \textup{direct answer}, \\
        \pi_{\theta}(o_{ 1: T} \mid x) \, \pi_{\theta}(o_{T+1:N} \mid x, o_{1:T}, I_{crop}), & \textup{tool call}, 
    \end{cases}
\end{aligned}
\end{equation}
where $N$ is the length of the entire generated sequence. In tool-call responses, $o_{1:T}$ represents \textit{tool tokens} in the first turn, and $o_{T+1:N}$ represents \textit{answer tokens} in the second turn, as illustrated in Fig.~\ref{fig:framework}. 
Let $n_{low}$ and $n_{crop}$ be the number of visual tokens for $I_{low}$ and $I_{crop}$. $\mathbf{1}_{\textup{tool}}$ is the indicator for tool-call responses. Thus, the total number of visual tokens for each sample is: $n_{img} = n_{low} + \mathbf{1}_{\textup{tool}}  n_{crop}$. 
Therefore, to minimize the number of visual tokens $n_{img}$, we aim to learn a policy $\pi_{\theta}(o \mid x)$ that can: 
(1) invoke the tool to request additional visual tokens only when necessary, and 
(2) acquire the minimal additional visual information $I_{crop}$ required to answer the question correctly. 


\subsection{Reward Design}

To learn a policy that can optimally balance efficiency and accuracy, we design a reward function that consists of two parts: (1) an Outcome Reward $\mathcal{R}_{oc}$ that reflects answer correctness, response format adherence and tool call frequency; (2) a Tool Reward $\mathcal{R}_{tool}$ that incentivizes effective tool exploration to enhance coarse-to-fine visual reasoning. 
The reward function of AdaptVision is:
\begin{equation}
    \mathcal{R} = \mathcal{R}_{oc} + \mathcal{R}_{tool}.
\end{equation}

\paragraph{Outcome Reward $\mathcal{R}_{oc}$.} The outcome reward is the sum of three components. 
(1) \textit{Accuracy reward} $\mathcal{R}_{acc}$: Since VQA answers are typically open-ended, we use an LLM as judge to assign a binary reward (1 for correct, 0 for incorrect) for answer correctness. 
(2) \textit{Format reward} $\mathcal{R}_{form}$: To maintain instruction-following capability, we enforce formatting requirements: reasoning in \texttt{<think>} tags, answers in \texttt{<answer>} tags, and tool calls in \texttt{<tool\_call>} tags with valid JSON. The format reward is 0.5 for full compliance with all formatting requirements; otherwise, the reward is 0.
(3) \textit{Balance reward} $\mathcal{R}_{bal}$: To prevent over-reliance on tool calls,
we introduce a balance reward. We apply a 0.1 penalty to correct answers that invoke tool calls. Additionally, to discourage ``lucky guesses''~\citep{yang2025visionthink}, we impose a 0.1 penalty on direct answers when the probability of correct response from low-resolution images is low, thereby encouraging appropriate tool usage. The design of this balance reward is as follows:
\begin{gather} \label{eq:R_bal}
    \mathcal{R}_{bal} = 
    \begin{cases}
        - 0.1 \cdot \mathbb{I} (r < \theta) \cdot \mathbb{I} (\mathcal{R}_{acc}=1), & \textup{direct answer}, \\
        - 0.1 \cdot \mathbb{I} (\mathcal{R}_{acc}=1), & \textup{tool call} ,
    \end{cases} 
    \\
    r = \frac{C_{direct}}{C_{direct} + C_{tool}},
\end{gather}
where $C_{direct}$ and $C_{tool}$ represent the count of correct answers for direct-answer and tool-call responses within a group, respectively. $\mathbb{I}$ is the indicator function. We set $\theta=0.2$ in this paper. 

\paragraph{Tool Reward $\mathcal{R}_{tool}$.} When the model requests additional visual information via a tool call, the cropped region $I_{crop}$ must be both informative for answering and minimal in area to reduce visual token usage. To achieve this balance, we introduce a tool reward $\mathcal{R}_{tool}$, formulated as follows:
\begin{equation} \label{eq:R_tool}
    \mathcal{R}_{tool} = \mathcal{R}_{crop} - \alpha \cdot \mathcal{R}_{area},
\end{equation}
where $\mathcal{R}_{crop}$ evaluates the correctness of the cropped region, $\mathcal{R}_{area}$ denotes its relative area ratio, and $\alpha$ is a hyperparameter balancing the two terms. In this paper we set $\alpha=2$. 
(1) The \textit{crop reward} $\mathcal{R}_{crop}$ is determined by GPT-4o, which evaluates whether the cropped region $I_{crop}$ contains relevant information to answer the question, returning 1 if correct and 0 otherwise. 
(2) The \textit{relative area reward} $\mathcal{R}_{area}$ penalizes oversized bounding boxes that contain irrelevant regions, formulated as follows:
\begin{gather}
\mathcal{R}_{area} = 
\mathbb{I}(\mathcal{R}_{acc} = 1) \cdot 
\mathbb{I}(\mathcal{R}_{crop} = 1) \cdot 
\text{clip} \left( 
    \frac{r_a}{\mu_a} - 1, 0, 1
\right), \notag
\\
r_a = \frac{(x_2 - x_1) \cdot (y_2 - y_1)}{H_{low} \cdot W_{low}},
\quad
\mu_a = \mu_{area}(\mathcal{G}(a)) ,
\end{gather}
where $H_{low}$ and $W_{low}$ denote the height and width of $I_{low}$, and $r_{a}$ is the area ratio of the cropped region. 
Here, $\mathcal{G}(a)$ denotes a group of responses that yield both correct answers ($\mathcal{R}_{acc}=1$) and correct cropped regions ($\mathcal{R}_{crop}=1$), and $\mu_{area}(\mathcal{G}(a))$ is the mean measurement of $r_{a}$ within such a group. This area penalty incentivizes the model to select the smallest possible region that still ensures correctness, thereby minimizing visual token usage while maintaining performance.
\subsection{Efficient Learning via Decoupled Turn Policy Optimization}

Based on our reward design, we initially employ GRPO~\citep{shao2024deepseekmath} for training. 
We aim to train a VLM that (1) achieves high answering accuracy and (2) minimizes the number of visual tokens used. 
However, training such a dual-objective policy with GRPO presents two key challenges.  

\paragraph{Ambiguous credit assignment} Vanilla GRPO provides a single, sequence-level reward to all generated tokens, failing to distinguish between the contributions of two distinct types of actions -- the decision to request additional visual tokens and the generation of the final answer. This ambiguity limits effective exploitation and exploration during policy learning. 
For instance, when the VLM correctly generates bounding boxes while producing an incorrect answer,
the model still receives a positive reward for the answer tokens. This may steer the model towards a suboptimal optimization direction. As we will show in the experiments, when training with GRPO, the model initially favors direct answers but then rapidly collapses to excessive tool call, resulting in an unstable training process. 


\paragraph{Imbalanced optimization} As defined in Eq.~\ref{eq:pi}, the policy model generates either a one-turn or two-turn responses for each sample. Depending on their functional roles, the generated tokens can be categorized into two types: \textit{Tool Tokens} and \textit{Answer Tokens}, as shown in Fig.~\ref{fig:framework}. 
Accordingly, the original GRPO objective in Eq.~\ref{eq:grpo} can be decomposed into two components that separately optimize the tool and answer tokens: 
\begin{multline} \label{eq:grpo_two_token}
    \frac{1}{G} \sum_{t=1}^{G} \frac{1}{N_i} \sum_{t=1}^{N_i} \mathcal{L}_{i,t}(\theta) = 
        \underbrace{\frac{1}{G} \sum_{t=1}^{G} \frac{1}{N_i} \sum_{t=1}^{T_i} \mathcal{L}_{i,t}(\theta) }_{\textup{Tool Token}} \\
        + \underbrace{\frac{1}{G} \sum_{t=1}^{G} \frac{1}{N_i} \sum_{t=T_i+1}^{N_i} \mathcal{L}_{i,t}(\theta)}_{\textup{Answer Token}},
\end{multline}
where $T_i$ denotes the number of tool tokens generated in the first turn, and $N_i - T_i$ represents the number of answer tokens in the second turn. If the model answers directly without tool calls, $T_i$ is 0. 
A closer examination of Eq.~\ref{eq:grpo_two_token} reveals an inherent optimization imbalance. 
In two-turn sequences that invoke tools, the gradient contributions from tool tokens are suppressed by the normalization factors $\frac{1}{N_i}$ and $\frac{1}{G}$, causing tool tokens to be under-optimized compared to answer tokens.

\begin{figure}[t]
    \begin{center}
        \includegraphics[width=\columnwidth]{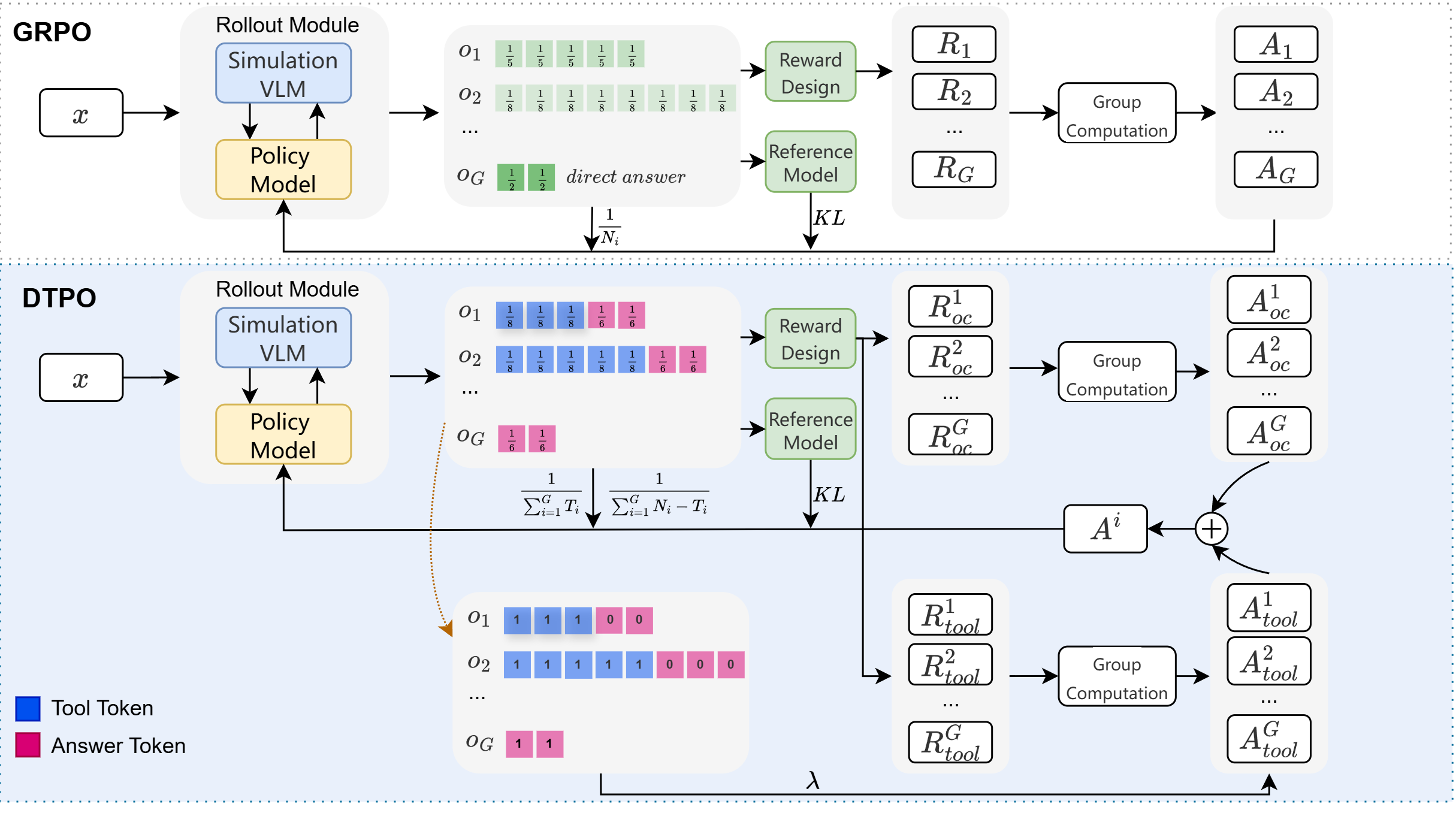}
    \end{center}
    \vspace*{-2mm}
    \caption{\textbf{Demonstration of vanilla GRPO and our DTPO.} Our DTPO (1) decomposes the policy loss by turns to separately optimize tool and answer tokens, and (2) computes distinct advantages for tool and outcome rewards, enabling balanced optimization and precise credit assignment. 
    }
    \label{fig:dtpo}
    \vskip -0.2in
\end{figure}

\begin{table*}[t!]
\centering
\caption{\textbf{Performance comparison with previous efficient VLM methods.} Vanilla denotes the Qwen2.5-VL-7B-Instruct model. Down-Sample uses a 1/4-resolution image as input to the Vanilla model. ``\#Token'' indicates the visual token consumption ratio relative to the vanilla model across all benchmarks. ``Avg.'' denotes the average performance relative to the vanilla model on all benchmarks.}

\resizebox{1\linewidth}{!}{
\begin{tabular}{@{}l|ccccccccc|cc}
\toprule
\multirow{2}{*}{\centering \textbf{Method}} & \textbf{ChartQA} & \textbf{OCRBench} & \textbf{DocVQA} & \textbf{MME} & \textbf{MMVet} & \textbf{RealWorldQA} & \textbf{POPE} & \textbf{MathVista} & \textbf{MathVerse} & \multirow{2}{*}{\centering \textbf{\#Token}$\downarrow$}& \multirow{2}{*}{\centering \textbf{Avg.}$\uparrow$} \\
& test & test & val & test & test & test & test & testmini & testmini & \\
\midrule
\rowcolor{gray!35}
\multicolumn{12}{c}{\textit{Retain 100\% Visual Tokens Across All Benchmarks}}\\
\multirow{2}{*}{\centering Vanilla} & 79.8 & 81.5 & 95.1 & 2316 & 61.6 & 68.6 & 86.7 & 68.2 & 46.3 & \multirow{2}{*}{\centering 100\%} & \multirow{2}{*}{\centering 100\%} \\
& 100\% & 100\% & 100\% & 100\% & 100\% & 100\% & 100\% & 100\% & 100\% \\
\midrule
\rowcolor{gray!35}
\multicolumn{12}{c}{\textit{Retain 25\% Visual Tokens Across All Benchmarks}}\\
\multirow{2}{*}{\centering Down-Sample} & 62.9 & 68.8 & 94.3 & 2270 & 54.5 & 68.8 & 82.8 & 62.2 & 43.1 & \multirow{2}{*}{\centering 25\%} & \multirow{2}{*}{\centering 92.1\%} \\
& 78.8\% & 84.4\% & 99.1\% & 98.0\% & 88.5\% & 100.3\% & 95.5\% & 91.2\% & 93.1\% \\
\midrule
\rowcolor{gray!35}
\multicolumn{12}{c}{\textit{Retain 50\% Visual Tokens Across All Benchmarks}}\\
\multirow{2}{*}{\centering SparseVLM} & 73.2 & 75.6 & 66.8 & 2282 & 51.5 & 68.4 & 85.5 & 66.6 & 45.1 & \multirow{2}{*}{\centering 50\%} & \multirow{2}{*}{\centering 92.2\%} \\
& 91.7\% & 92.7\% & 70.2\% & 98.5\% & 83.6\% & 99.7\% & 98.6\% & 97.6\% & 97.4\% \\
\midrule
\multirow{2}{*}{\centering FastV} & 72.6 & 75.8 & 93.6 & 2308 & 52.8 & 68.8 & 84.7 & 63.7 & 45.0 & \multirow{2}{*}{\centering 50\%} & \multirow{2}{*}{\centering 95.8\%} \\
& 91.0\% & 93.0\% & 98.4\% & 99.6\% & 85.7\% & 100.3\% & 97.7\% & 93.4\% & 97.2\% \\
\midrule
\multirow{2}{*}{\centering VisionZip} & 71.5 & 70.5 & 93.8 & 2209 & 57.0 & 68.6 & 86.3 & 64.1 & 45.1 & \multirow{2}{*}{\centering 50\%} & \multirow{2}{*}{\centering 94.8\%} \\
& 89.6\% & 86.5\% & 98.6\% & 95.4\% & 92.5\% & 100\% & 99.5\% & 93.9\% & 97.4\% \\
\midrule
\rowcolor{gray!35}
\multicolumn{12}{c}{\textit{Dynamic Methods}}\\
\multirow{2}{*}{\centering VisionThink} & 73.6 & 76.8 & 92.9 & 2320 & 61.7 & 65.6 & 86.3 & 62.2 & 42.5 & \multirow{2}{*}{\centering 52\%} & \multirow{2}{*}{\centering 95.8\%} \\
& 92.2\% & 94.2\% & 97.7\% & 100.2\% & 100.2\% & 95.6\% & 99.5\% & 91.2\% & 91.8\% \\
\midrule
\multirow{2}{*}{\centering VisionThink\textsuperscript{\textdagger}} & 73.88 & 80.8 & 93.7 & 2392 & 60.18 & 68.37 & 86.69 & 65.7 & 45.68 & \multirow{2}{*}{\centering 99\%} & \multirow{2}{*}{\centering 98.4\%} \\
& 92.6\% & 99.1\% & 98.5\% & 103.3\% & 97.7\% & 99.7\% & 100.0\% & 96.3\% & 98.7\% \\
\midrule
\multirow{2}{*}{\centering AdaptVision {\small w/o DTPO}} & 73.74 & 75.9 & 93.1 & 2354 & 61.28 & 65.7 & 86.8 & 64.4 & 44.2 & \multirow{2}{*}{\centering {57\%}} & \multirow{2}{*}{\centering {96.7\%}} \\
& 92.4\% & 93.1\% & 97.9\% & 101.6\% & 99.5\% & 95.8\% & 100.1\% & 94.4\% & 95.5\% \\
\midrule
\multirow{2}{*}{\centering AdaptVision} & 75.92 & 76.9 & 92.6 & 2379 & 64.8 & 67.32 & 86.8 & 65.9 & 42.3 & \multirow{2}{*}{\centering {\textbf{33\%}}} & \multirow{2}{*}{\centering {\textbf{97.9\%}}} \\
& 95.1\% & 94.4\% & 97.4\% & 102.7\% & 105.2\% & 98.1\% & 100.1\% & 96.6\% & 91.4\% \\
\bottomrule
\end{tabular}}
\label{table:main_result}
\end{table*}

To address these challenges, we propose Decoupled Turn Policy Optimization (DTPO). 
First, 
we decouple the policy loss by turns and normalize the contributions of tool and answer tokens separately. This adjustment effectively resolves the under-optimization problem of tool tokens. 
\begin{multline}
    \mathcal{J}_{\text{DTPO}}(\theta) = \mathbb{E}_{x, o_i} 
    \Bigg[
            \underbrace{\frac{1}{\sum_{i=1}^G T_i} \sum_{i=1}^G  \sum_{t=1}^{T_i} \mathcal{L}_{i,t}(\theta)}_{\textup{Tool Token}} \\
            + \underbrace{\frac{1}{\sum_{i=1}^G (N_i - T_i)} \sum_{i=1}^G  \sum_{t=T_i+1}^{N_i} \mathcal{L}_{i,t}(\theta) }_{\textup{Answer Token}}
    \Bigg].
\end{multline}
Second, to enable precise credit assignment, DTPO decouples the advantage estimation by computing distinct advantages for tool and answer tokens, rather than using a single advantage for the entire sequence. Specifically, we compute the advantage for the $t$-th token as follows:
\begin{gather}
    A_{i,t} = 
    \begin{cases}
        A_{oc}^{(i)} + \lambda \cdot A_{tool}^{(i)}, & \textup{direct answer},  \\ 
        A_{oc}^{(i)} + \lambda \cdot A_{tool}^{(i)} \cdot \mathbb{I}(1 \le t \le T_{i}) , & \textup{tool call},
    \end{cases} \notag
    \\
    A_{tool}^{(i)} = \frac{\mathcal{R}_{tool}^{(i)} - \text{mean}(\{\mathcal{R}_{tool}^{(i)}\}^G_{i=1})}{\text{std}(\{\mathcal{R}_{tool}^{(i)}\}^G_{i=1})}, 
    \\
    A_{oc}^{(i)} = \frac{\mathcal{R}_{oc}^{(i)} - \text{mean}(\{\mathcal{R}_{oc}^{(i)}\}^G_{i=1})}{\text{std}(\{\mathcal{R}_{oc}^{(i)}\}^G_{i=1})}, 
\end{gather}
where $A_{tool}^{(i)}$ and $A_{oc}^{(i)}$ are advantage estimates computed using tool reward and outcome reward respectively, and $\lambda$ is a hyperparameter that trade-offs two advantages. We set $\lambda=0.3$ in this paper. Fig.~\ref{fig:dtpo} compares the design of GRPO and DTPO.

\section{Experiment}

\subsection{Evaluation Setup} \label{exp_setup}
We conduct experiments on several general VQA benchmarks, including ChartQA~\citep{masry2022chartqa}, OCRBench~\citep{liu2024ocrbench}, DocVQA~\citep{mathew2021docvqa}, MME~\citep{fu2024mmecomprehensiveevaluationbenchmark}, MMVet~\citep{yu2023mm}, RealWorldQA~\citep{xai2024grok1.5v_online}, POPE~\citep{li2023evaluatingobjecthallucinationlarge}, MathVista~\citep{lu2023mathvista},  MathVerse~\citep{zhang2024mathverse}. 
AdaptVision is based on Qwen2.5-VL-7B-Instruct~\citep{bai2025qwen2}. We employ veRL~\citep{sheng2025hybridflow} framework for RL training. During training, we set the batch size as 512 and the mini-batch size as 32. We drop the KL term during policy optimization. The initial learning rate of the policy model is $1e-6$. For each prompt, we sample 16 candidate responses using a temperature of 1.0. During inference, we use the vLLM framework and set the temperature to 0. We use training data from \citet{yang2025visionthink}\footnote{https://huggingface.co/datasets/Senqiao/VisionThink-Smart-Train}, which contains VQA samples that can be answered directly using low-resolution images, as well as samples that require high-resolution images for accurate answering. 

\subsection{Main Results}





\begin{figure}[t]
    \begin{center}
        \includegraphics[width=\columnwidth]{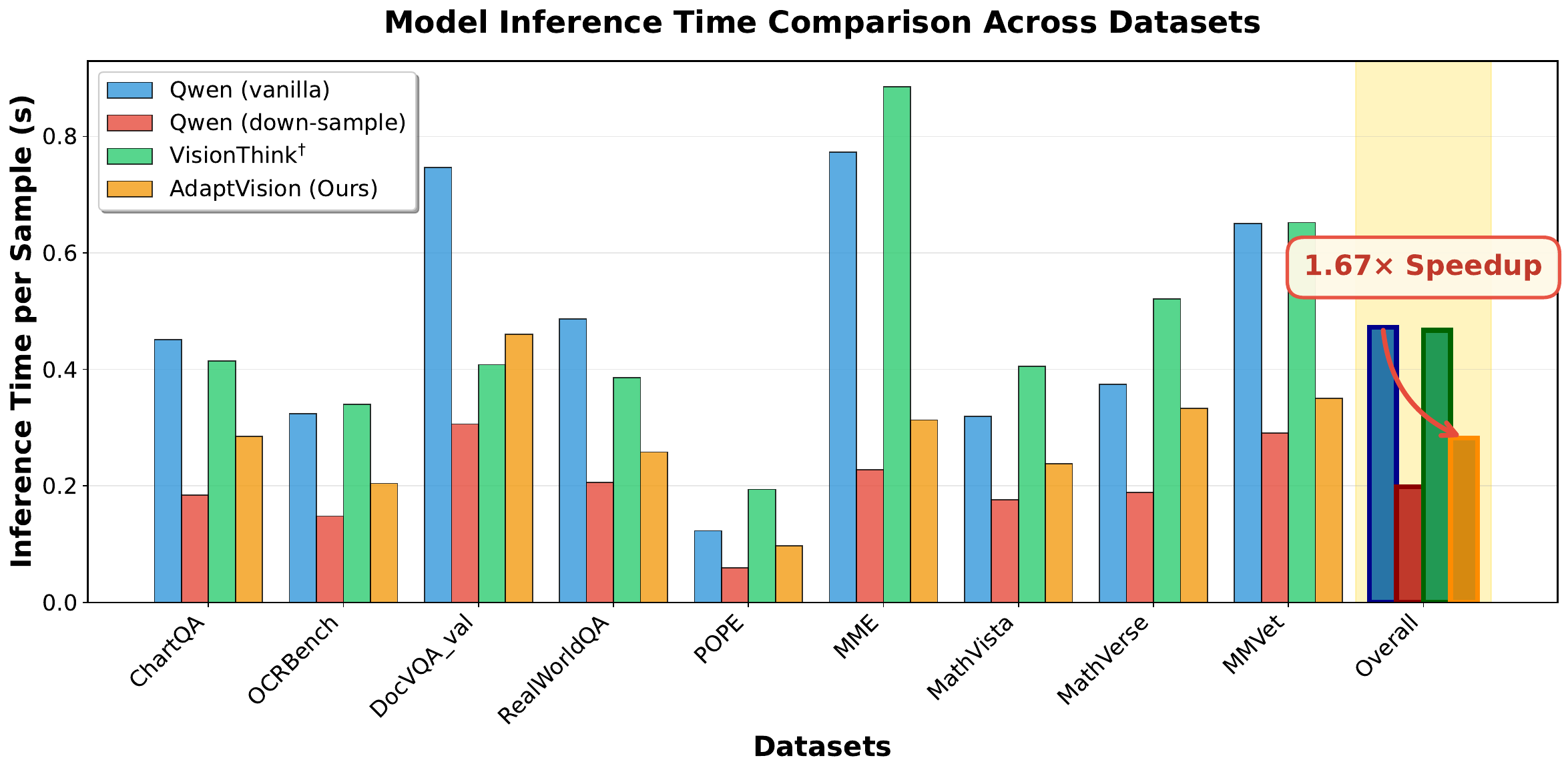}
    \end{center}
    \vspace*{-4mm} 
    \caption{\textbf{Comparison of Inference Time.} (1) Compared to the vanilla model and VisionThink\textsuperscript{\textdagger}, AdaptVision demonstrates significantly reduced inference time due to reduced visual token usage. (2) While AdaptVision requires additional generated tokens for reasoning and tool calls compared to the down-sample model, the resulting increase in inference time remains acceptable.}
    \label{fig:inference_time}
\end{figure}

\begin{figure*}[t!]
    \centering
    \begin{subfigure}[b]{0.242\linewidth}
        \centering
        \includegraphics[width=\linewidth]{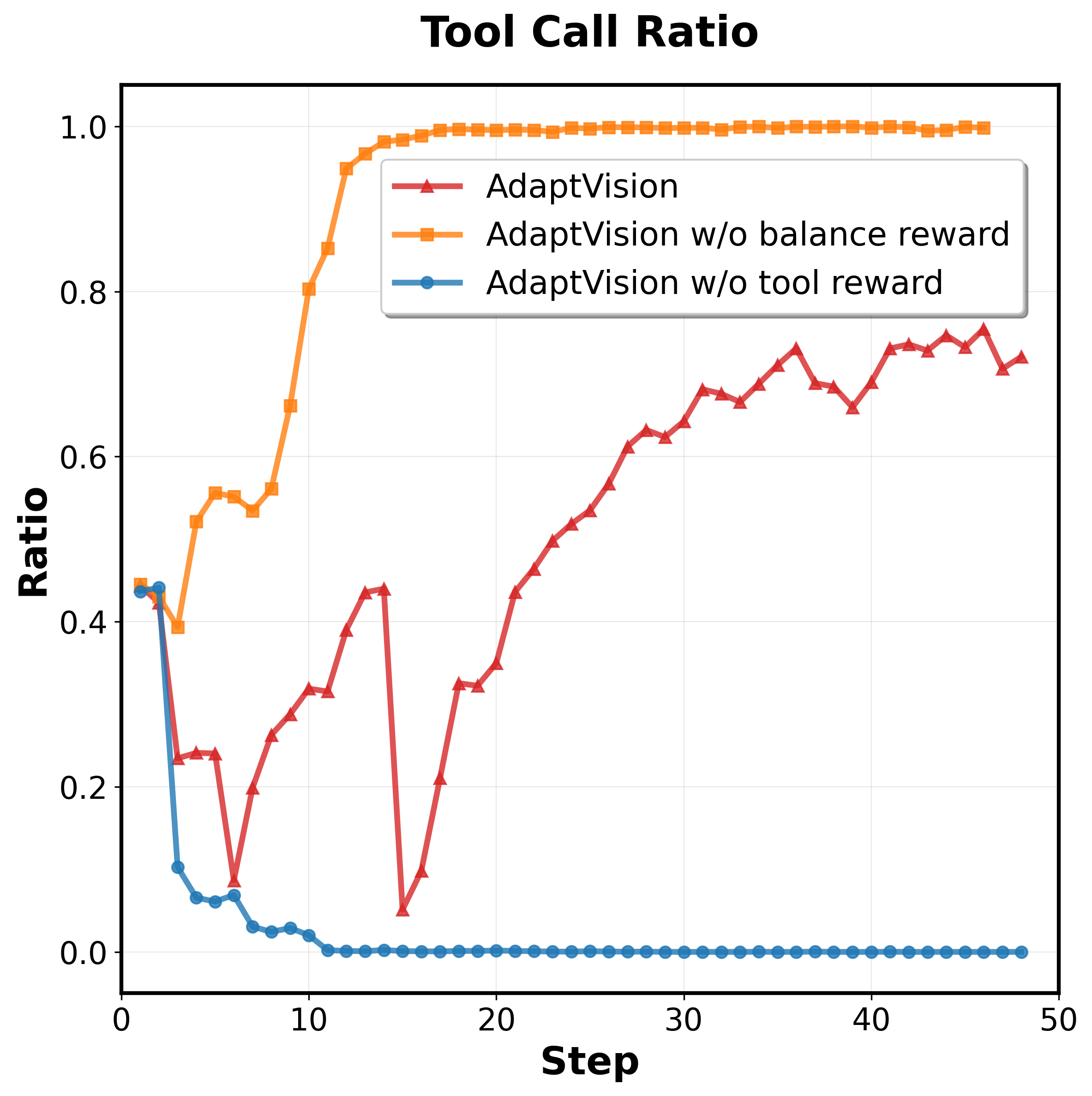}
        \caption{Reward Ablation}
        \label{fig:reward_ablation}
    \end{subfigure}
    \hfill
    \vrule
    \hfill
    \begin{subfigure}[b]{0.745\linewidth}        
        \centering
        \begin{subfigure}[b]{0.32\linewidth}
            \includegraphics[width=\linewidth]{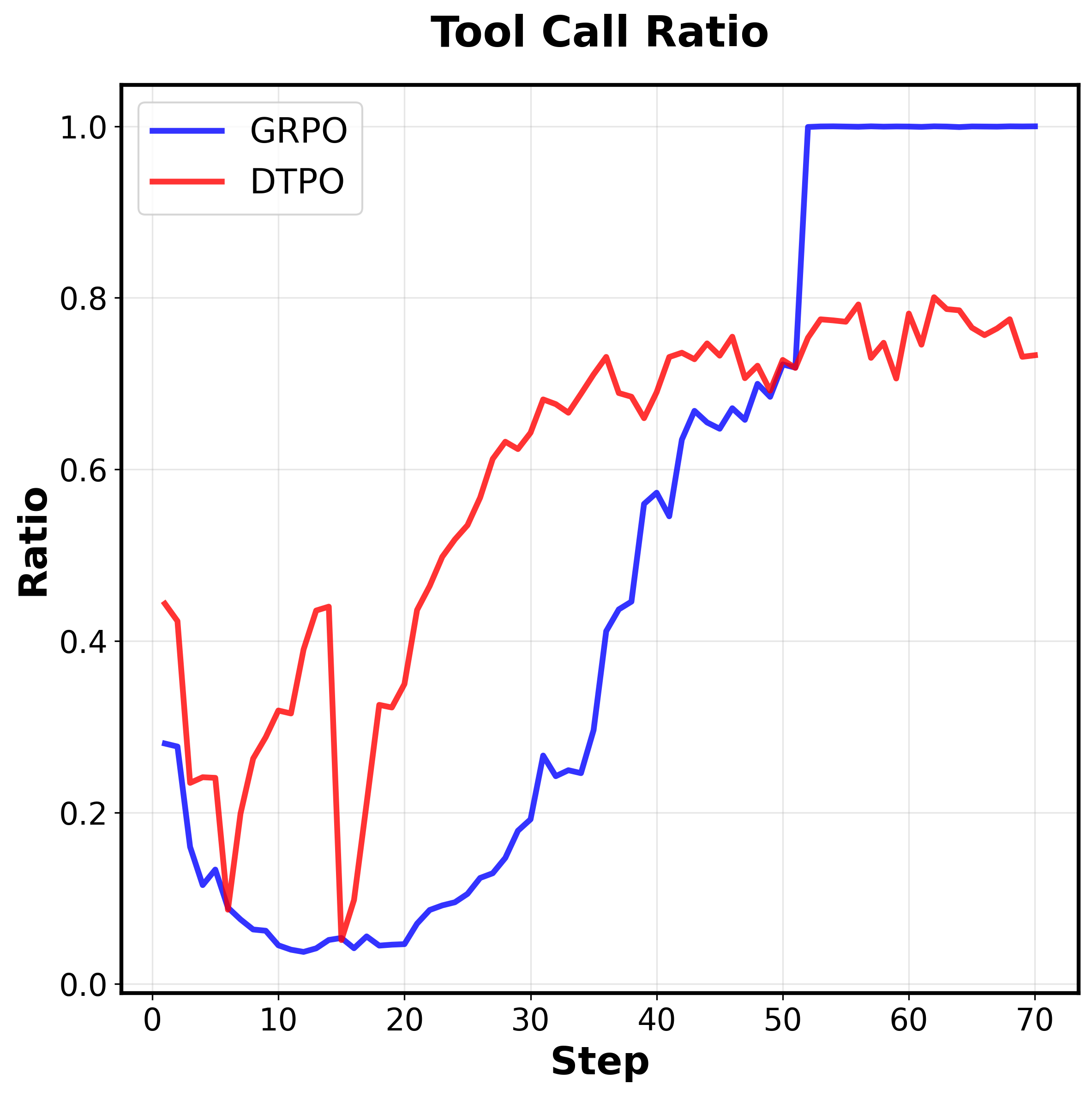}
        \end{subfigure}
        \hfill
        \begin{subfigure}[b]{0.32\linewidth}
            \includegraphics[width=\linewidth]{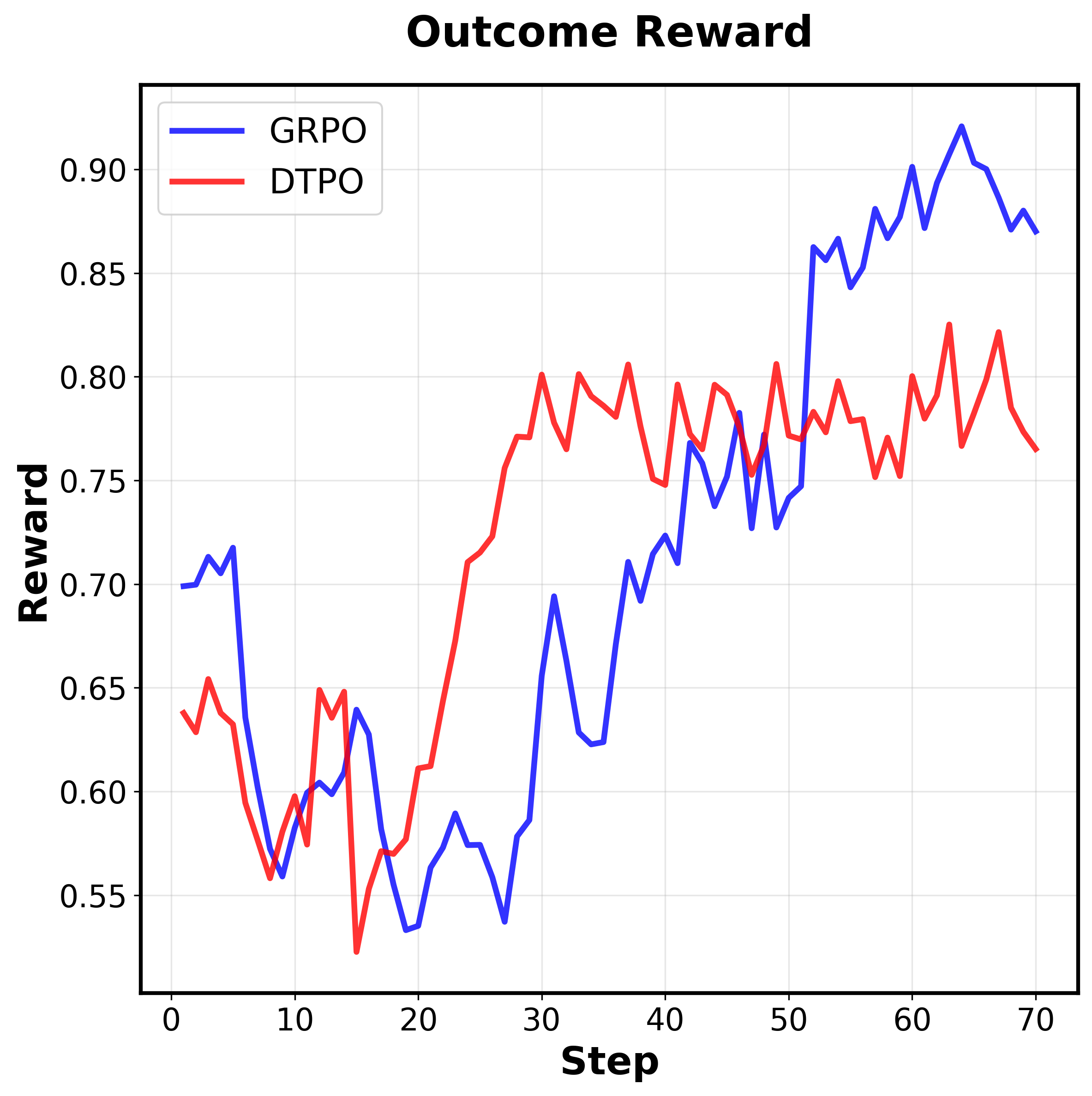}
        \end{subfigure}
        \hfill
        \begin{subfigure}[b]{0.32\linewidth}
            \includegraphics[width=\linewidth]{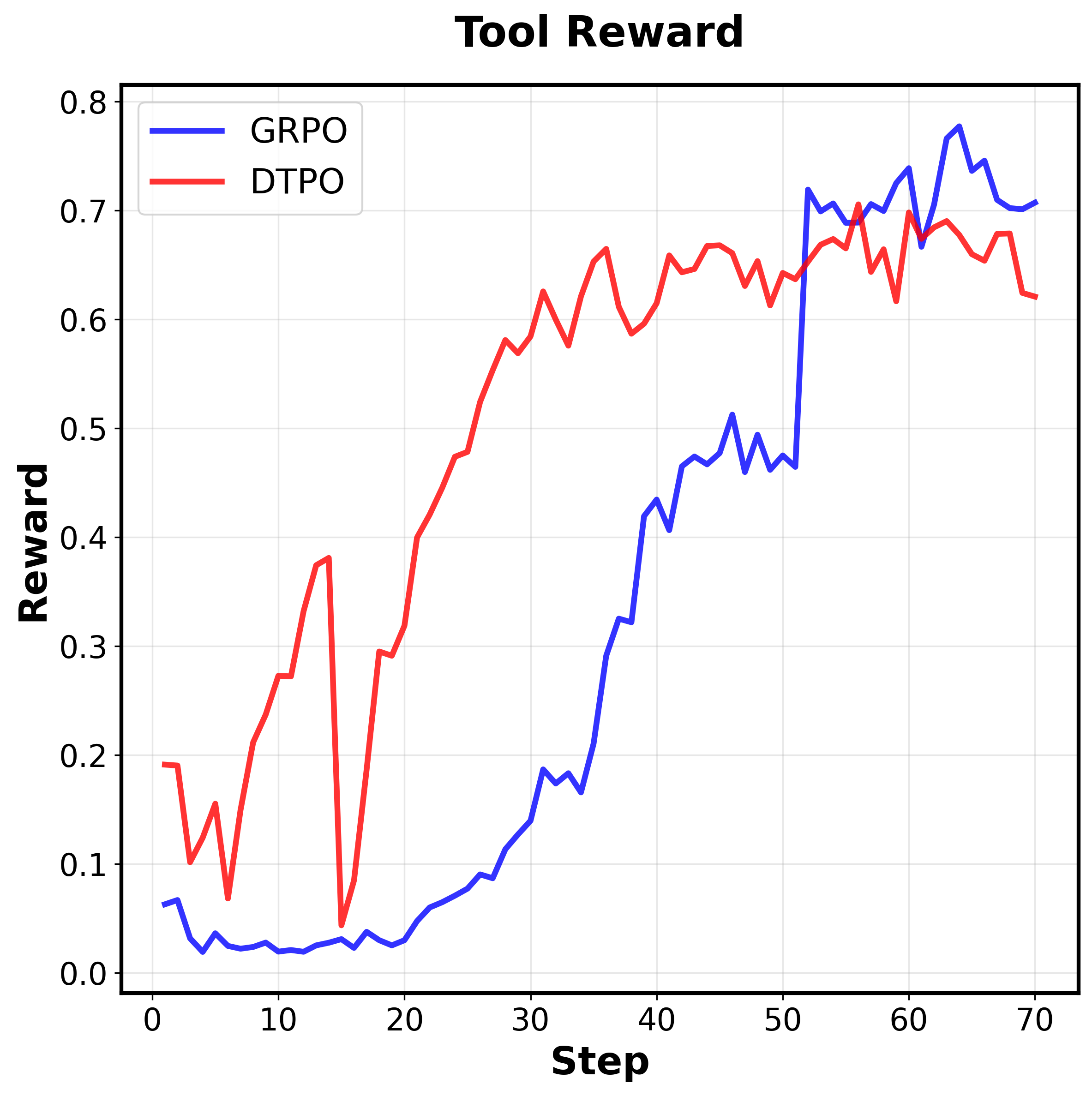}
        \end{subfigure}

        \caption{Training curve of tool call ratio, outcome reward and tool reward}
        \label{fig:training_process_grpo_vs_dtpo}
    \end{subfigure}
    \vspace{1mm}
    \caption{\textbf{Policy-training comparison}: (a) The influence of reward design. (b) GRPO vs. DTPO.}
    \label{fig:training_process}
\end{figure*}

\begin{figure*}[t]
    \centering
    \begin{subfigure}[b]{0.49\linewidth}        
        \centering
        \begin{subfigure}[b]{0.485\linewidth}
            \includegraphics[width=\linewidth]{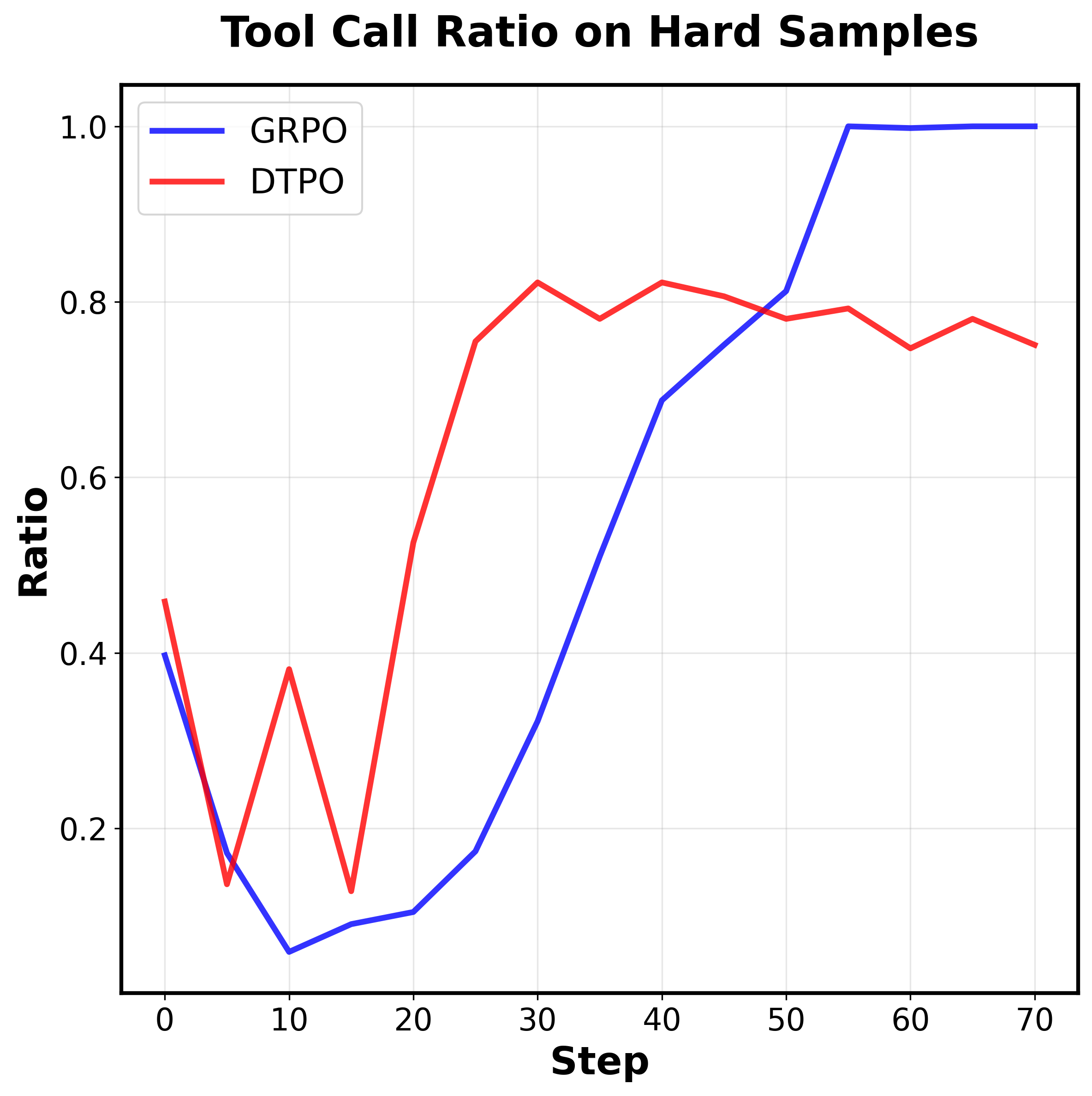}
        \end{subfigure}
        \hfill
        \begin{subfigure}[b]{0.485\linewidth}
            \includegraphics[width=\linewidth]{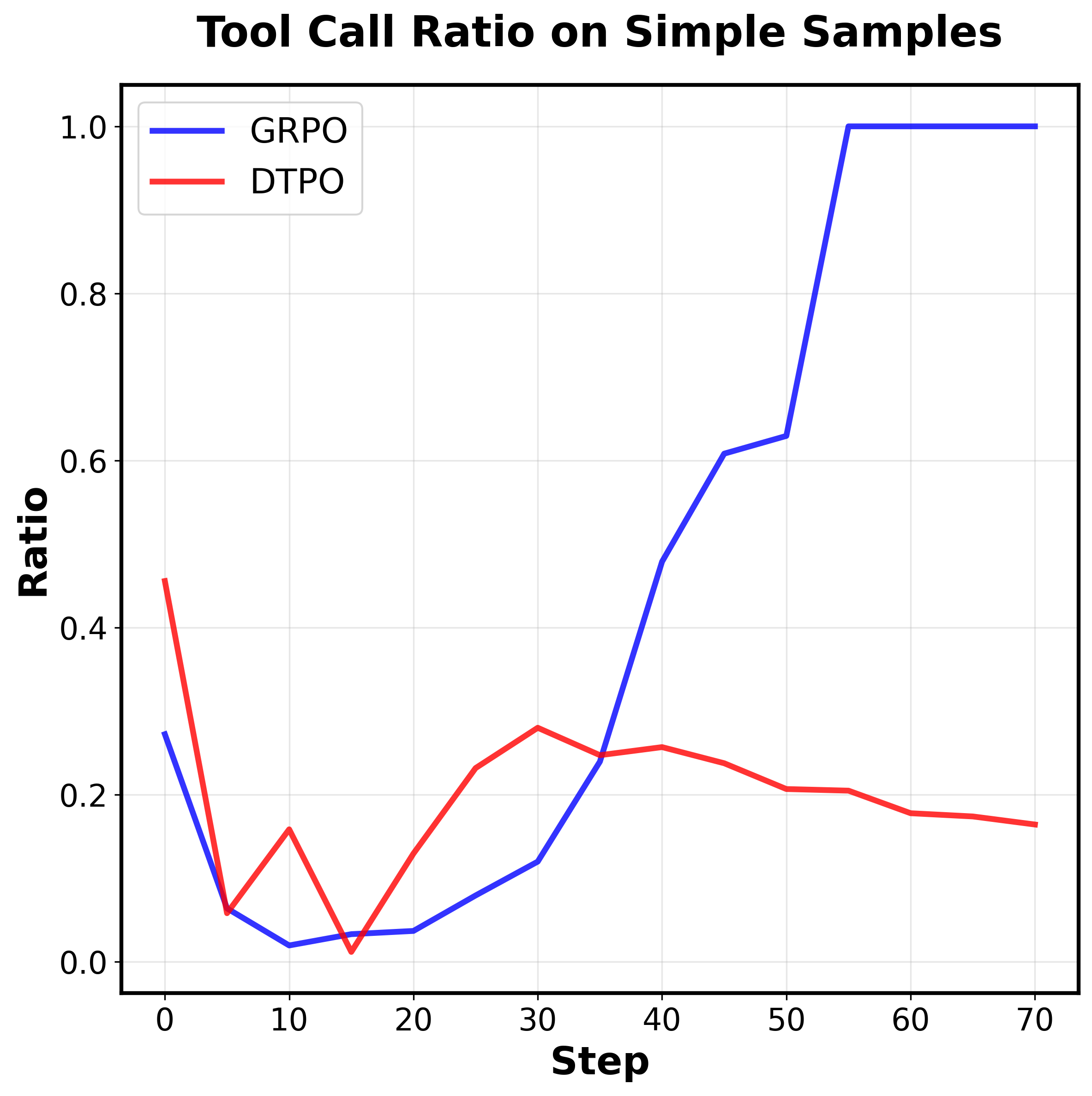}
        \end{subfigure}
    \caption{Training curve of tool call ratio on different types of data.}
    \label{fig:tool_call_different_data}
    \end{subfigure}
    \hfill
    \vrule
    \hfill
    \begin{subfigure}[b]{0.467\linewidth}        
        \centering
        \includegraphics[width=\linewidth]{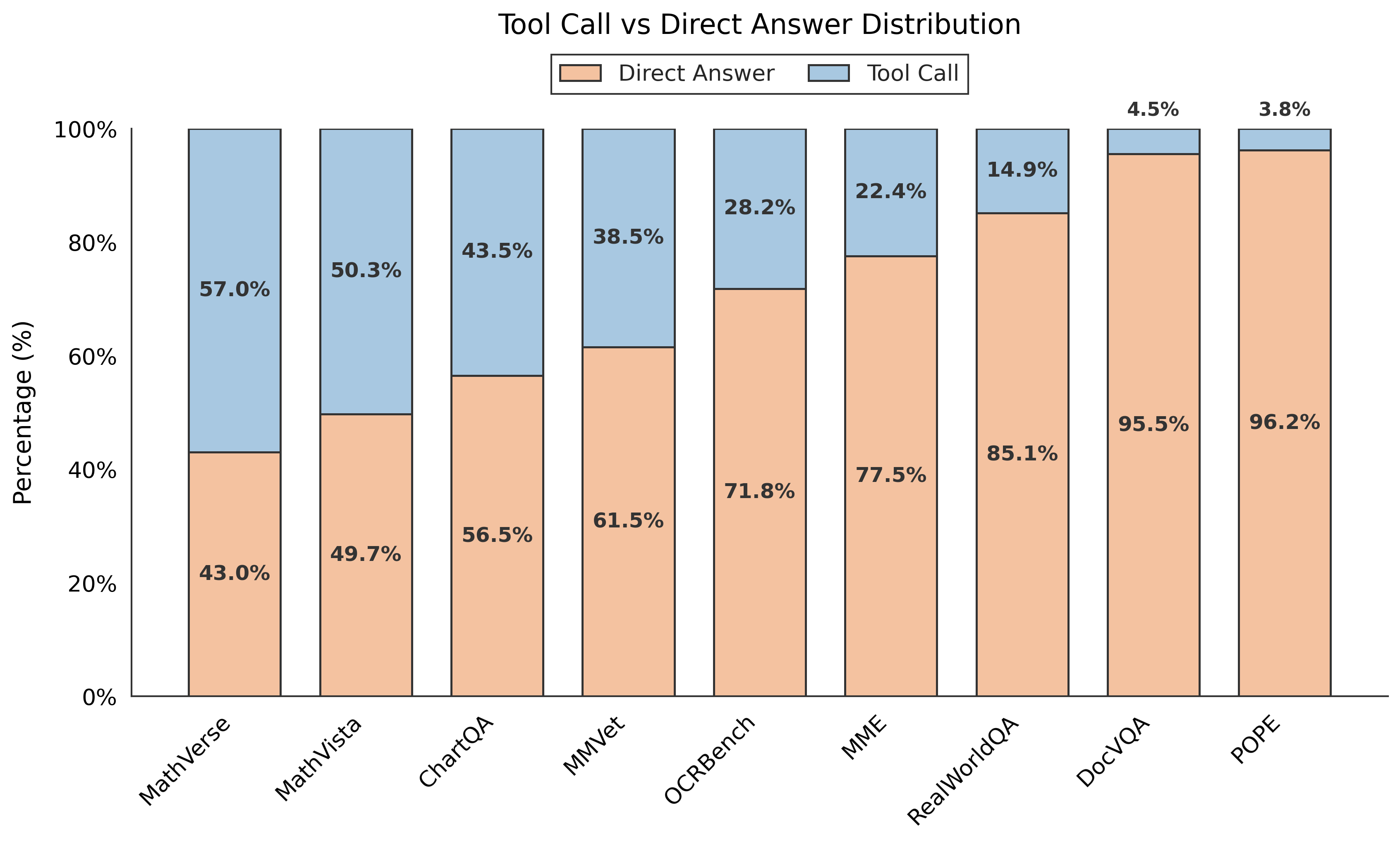}
        \caption{Tool call proportion across different benchmarks.}
        \label{fig:tool_call_ratio}
    \end{subfigure}
    \caption{\textbf{Tool call ratio analysis:} (a) Training curves show that DTPO learns a stable and adaptive policy, increasing tool calls for hard samples while decreasing them for simple ones. (b) Across different benchmarks, AdaptVision demonstrates a well-balanced ability to invoke tools when necessary and answer directly when possible.}
    \vskip -0.1in
\end{figure*}


We compare AdaptVision with existing vision token compression methods, including FastV~\citep{chen2024image}, SparseVLM~\citep{zhang2024sparsevlm}, VisionZip~\citep{yang2025visionzip}, and VisionThink~\citep{yang2025visionthink}. FastV, SparseVLM, and VisionZip are static methods that operate with a pre-defined token retention ratio, while VisionThink and AdaptVision are dynamic methods that vary visual token usage for each sample.
For fair comparison, static methods are set to 50\% token retention. 
For VisionThink, we initially used the officially released model\footnote{https://huggingface.co/Senqiao/VisionThink-Efficient} but found it consumed substantially more visual tokens than our method, making the comparison unfair. We thus report two versions: ``VisionThink\textsuperscript{\textdagger}'' for the released model and ``VisionThink'' for our reproduction using the public code\footnote{https://github.com/dvlab-research/VisionThink}. 
We also include the vanilla model (100\% tokens, high-resolution) and the down-sample model (25\% tokens, 1/4 resolution) based on Qwen2.5-VL-7B-Instruct. 
Results are shown in Table~\ref{table:main_result}. 
Compared to previous vision token compression methods (including FastV, SparseVLM, VisionZip and VisionThink), AdaptVision achieves superior average performance across all benchmarks with significantly fewer visual tokens. 
Compared to the down-sample model, AdaptVision improves accuracy by 5.8\% (92.1\% $\rightarrow$ 97.9\%) with only 7\% (25\% $\rightarrow$ 33\%) more visual tokens, highlighting its effective coarse-to-fine visual reasoning. 


\paragraph{Inference Latency} We also compare inference time across multiple models: the vanilla model, the down-sample model, and VisionThink\textsuperscript{\textdagger}. The end-to-end inference time measurements for each dataset are presented in Fig.~\ref{fig:inference_time}. AdaptVision demonstrates significantly reduced inference time (1.67x overall speedup) compared to both the vanilla model and VisionThink\textsuperscript{\textdagger}, primarily due to its more efficient visual token usage. While AdaptVision does require additional token generation for reasoning and tool calls compared to the down-sample model, the resulting increase in inference time remains within acceptable bounds.



\begin{figure*}[t]
    \centering
    \begin{subfigure}[b]{1\linewidth}        
        \centering
        \begin{subfigure}[b]{0.49\linewidth}
            \includegraphics[width=\linewidth]{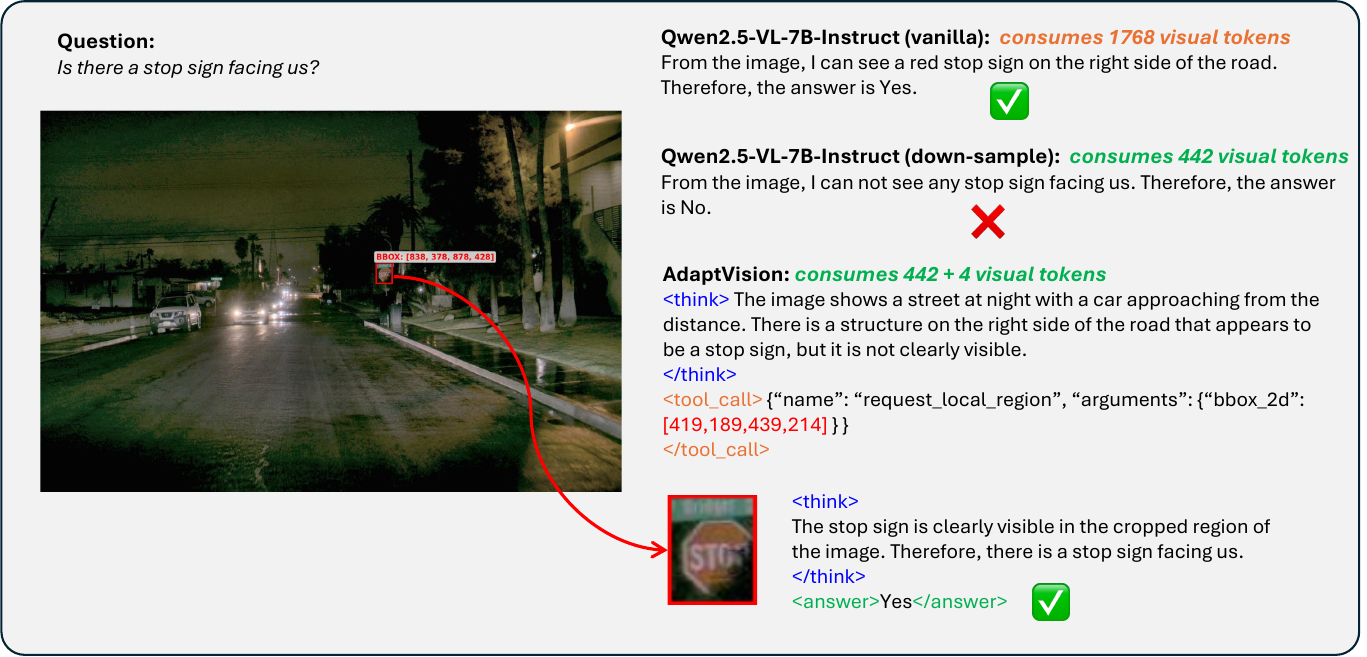}
        \end{subfigure}
        \hfill
        \begin{subfigure}[b]{0.49\linewidth}
            \includegraphics[width=\linewidth]{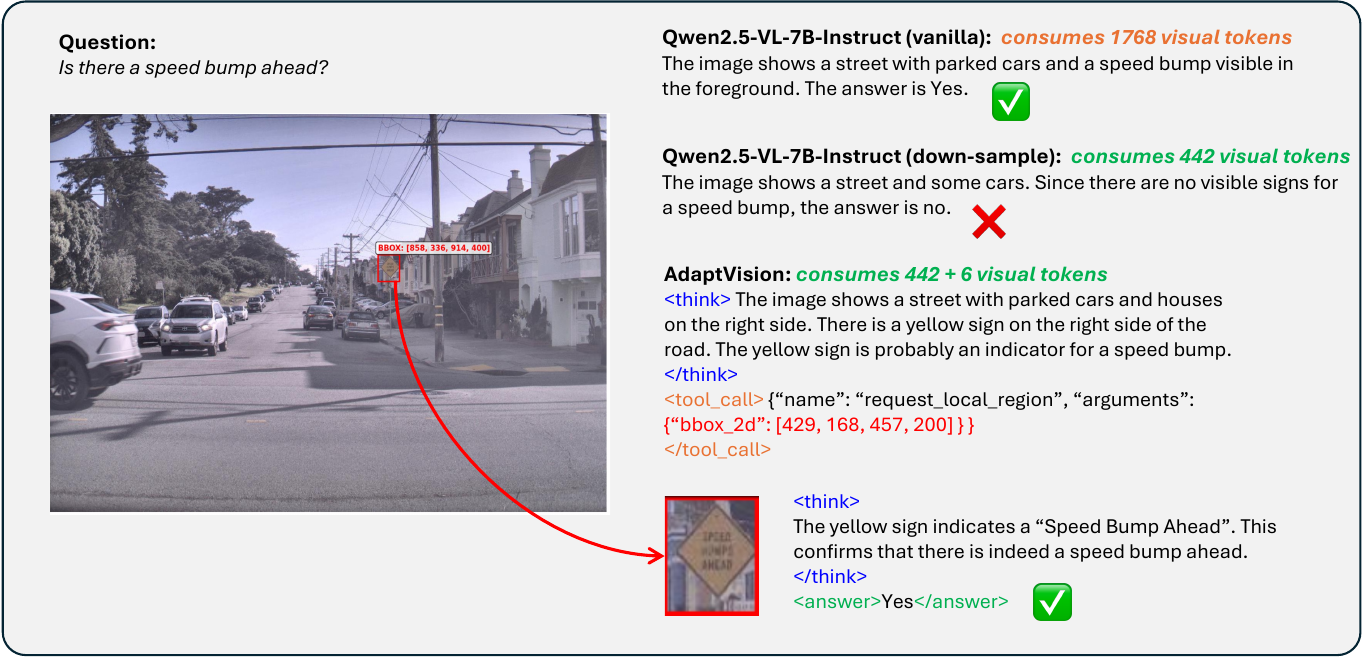}
        \end{subfigure}
    \end{subfigure}
    \caption{\textbf{Case study:} (1) The vanilla model yields a correct answer but consumes a large number of visual tokens; (2) The down-sample model reduces token usage but fails to answer correctly; (3) AdaptVision smartly invokes the tool to produce a correct answer with minimal visual token cost.}
    \label{fig:case1}
    \vskip -0.1in
\end{figure*}

\subsection{Analysis}

\paragraph{Reward Design}
To investigate the impact of reward design on model behavior, we conduct an ablation study on balance and tool rewards. As shown in Fig.~\ref{fig:reward_ablation}, the absence of the balance reward causes the model to quickly collapse to excessive tool use. This occurs because the tool reward incentivizes correct tool use, which generally improves accuracy as training progresses. 
Conversely, with balance reward, the VLM learns to adaptively regulate tool usage based on the input. 
Furthermore, the ablation of the tool reward reveals its necessity for exploration: without it, the model collapses to direct answering and fails to invoke the tool after just 10 training steps. In contrast, with the tool reward, the model successfully explores and leverages the tool to enhance performance.
\paragraph{GRPO vs. DTPO}
We compare the training processes of GRPO and DTPO in Fig.~\ref{fig:training_process_grpo_vs_dtpo}. GRPO exhibits an unstable training dynamic: During the early training phase, it struggles to optimize either the tool or outcome reward, causing the tool call ratio to drop near zero and limiting exploration. 
After approximately 20 steps, both rewards and the tool call ratio surge rapidly, shifting the model from direct answering to excessive tool use, eventually collapsing to tool call. This instability stems from GRPO's ambiguous credit assignment and imbalanced optimization. 
In contrast, DTPO exhibits a stable and efficient optimization process. Both rewards rise steadily from the start, reflecting effective tool use exploration. The model subsequently converges to a reasonable tool call ratio, demonstrating the effectiveness of DTPO. 
In Table~\ref{table:main_result}, we observe that the GRPO-trained model not only performs worse than DTPO, but it also uses far more visual tokens. This confirms that DTPO is critical for effectively learning the balance between tool use and direct answering.
Furthermore, we compare the tool call ratios across different data types. Fig.~\ref{fig:tool_call_different_data} illustrates that our model learns to selectively invoke tools based on task difficulty, while the model trained with GRPO calls tools on all samples, resulting in a 100\% tool call ratio.

\paragraph{Adaptive Tool-use}
We further investigate tool-use efficiency by measuring the proportion of tool call responses across various benchmarks. As shown in Fig.~\ref{fig:tool_call_ratio}, for complex visual tasks like MathVerse and ChartQA that require fine-grained details, the model frequently invokes the tool to better answer the question. For general understanding tasks like POPE, our model rarely calls the tool, thereby maintaining high efficiency. This shows our model has learned adaptive reasoning: it solves problems directly when tools are unnecessary while still leveraging them when beneficial.


\subsection{Case Study}

In this section, we present a case study to illustrate the efficient visual reasoning process of AdaptVision. 
We compare AdaptVision with the vanilla model and the down-sample model.
As shown in Fig.~\ref{fig:case1}, the down-sample model, while reducing visual token usage, fails to answer correctly due to insufficient information in the low-resolution image. 
The vanilla model, using the original high-resolution image, yields a correct answer but at the cost of a large number of visual tokens. 
In contrast, AdaptVision begins with the low-resolution image, analyzes the question and image, recognizes the informational inadequacy, and then intelligently invokes the tool to crop the most relevant region from the high-resolution image. By acquiring only this essential additional visual information, it produces an accurate answer while minimizing visual token consumption. 


\section{Conclusion}

In this paper, we present AdaptVision, a novel paradigm that enables VLMs to autonomously determine the minimum number of visual tokens via adaptive, coarse-to-fine visual reasoning. 
We propose a Decoupled Turn Policy Optimization (DTPO) algorithm, which handles dual-objective policy learning by decoupling the learning objective and advantage estimation. This leads to a more balanced and effective training process than GRPO.
Experiments on multiple VQA benchmarks show that AdaptVision achieves superior performance using significantly fewer visual tokens than previous efficient VLM methods. These results advance the development of computationally efficient and biologically inspired VLMs. 


Despite its effectiveness, AdaptVision has several limitations that outline directions for future research. First, our framework currently relies on a single visual tool and a fixed initial compression ratio (1/4 resolution). Expanding the toolset and enabling dynamic resolution selection could further enhance adaptability. Second, the reasoning process is constrained to a maximum of two turns, which may limit deep visual reasoning for complex tasks. We hope that future research will address these limitations, further advancing the development of efficient VLMs.



{
    \small
    \bibliographystyle{ieeenat_fullname}
    \bibliography{main}
}
\clearpage
\appendix
\section{Additional Details} \label{app:add_exp}

\subsection{Prompt Details} \label{app:prompt_detail}

AdaptVision utilizes three types of prompts. First, to equip the VLM with basic tool-using capability, we follow the Qwen2.5-VL cookbook~\citep{bai2025qwen2} to design prompts for the bounding box tool (Table~\ref{tab:vlm_prompt}). 
Second, since VQA tasks are typically diverse and open-ended, we adopt an LLM-as-judge approach to evaluate answer correctness. As shown in Table~\ref{tab:llm_as_answer_judge}, following \citet{yang2025visionthink}, we design a judging prompt for GPT-4o to produce binary evaluations (1 for correct, 0 for incorrect). 
Third, to encourage efficient tool exploration, we prompt GPT-4o to evaluate the relevance of cropped regions, producing a binary reward for region correctness (Table~\ref{tab:judge_bbox_correctness}).

\subsection{Training and Evaluation Details} \label{app:training_detail}
AdaptVision is based on Qwen2.5-VL-7B-Instruct~\citep{bai2025qwen2}. We employ veRL~\citep{sheng2025hybridflow} framework for RL training. During training, we set the batch size as 512 with mixed-precision (FP16) training. The mini-batch size is 32. We drop the KL term during policy optimization. 
For each prompt, we sample 16 candidate responses (i.e., $G=16$) using a temperature of 1.0. The upper and lower clip ratios are 0.24 and 0.20, respectively. We set the maximum prompt length and the maximum response length as 8192. All experiments were conducted on 4 nodes, each with 8 H20 GPUs. 
The model was trained for 80 steps, using the AdamW optimizer with a learning rate of $1e-6$, $\beta=(0.9, 0.999)$, and a weight decay of 0.01.
During inference, we use the vLLM framework and set the temperature to 0. 

\subsection{Additional Results} \label{app:more_results}
We further compare AdaptVision with previous efficient VLM methods with different visual token retention ratios. As shown in Table~\ref{table:more_result}, while the performance of FastV, SparseVLM, and VisionZip degrades with reduced token ratios, AdaptVision maintains superior performance with significantly fewer visual tokens. 


\section{Qualitative Results} \label{app:qualitative_results}

\begin{figure}[ht]
    \begin{center}
        \includegraphics[width=\linewidth]{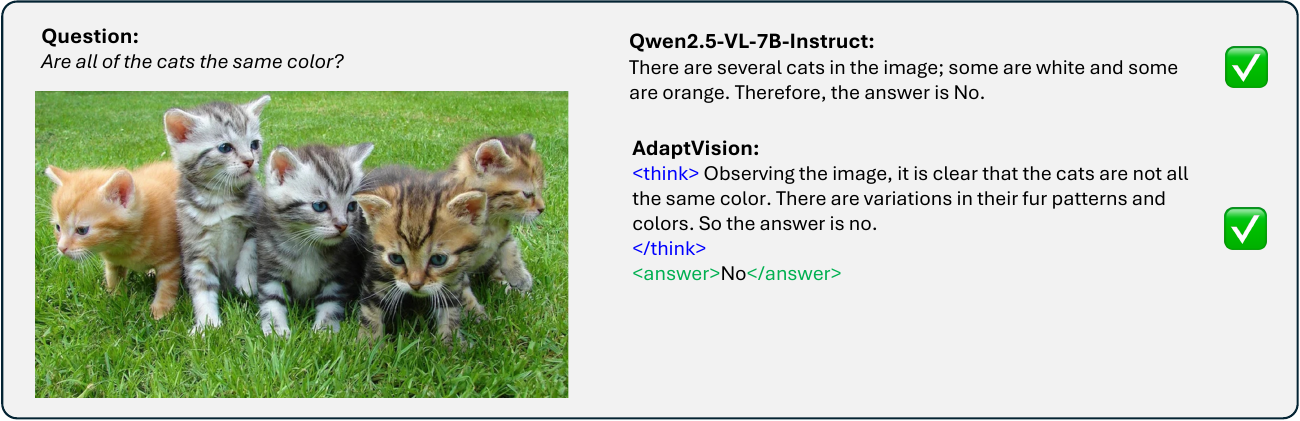}
        \includegraphics[width=\linewidth]{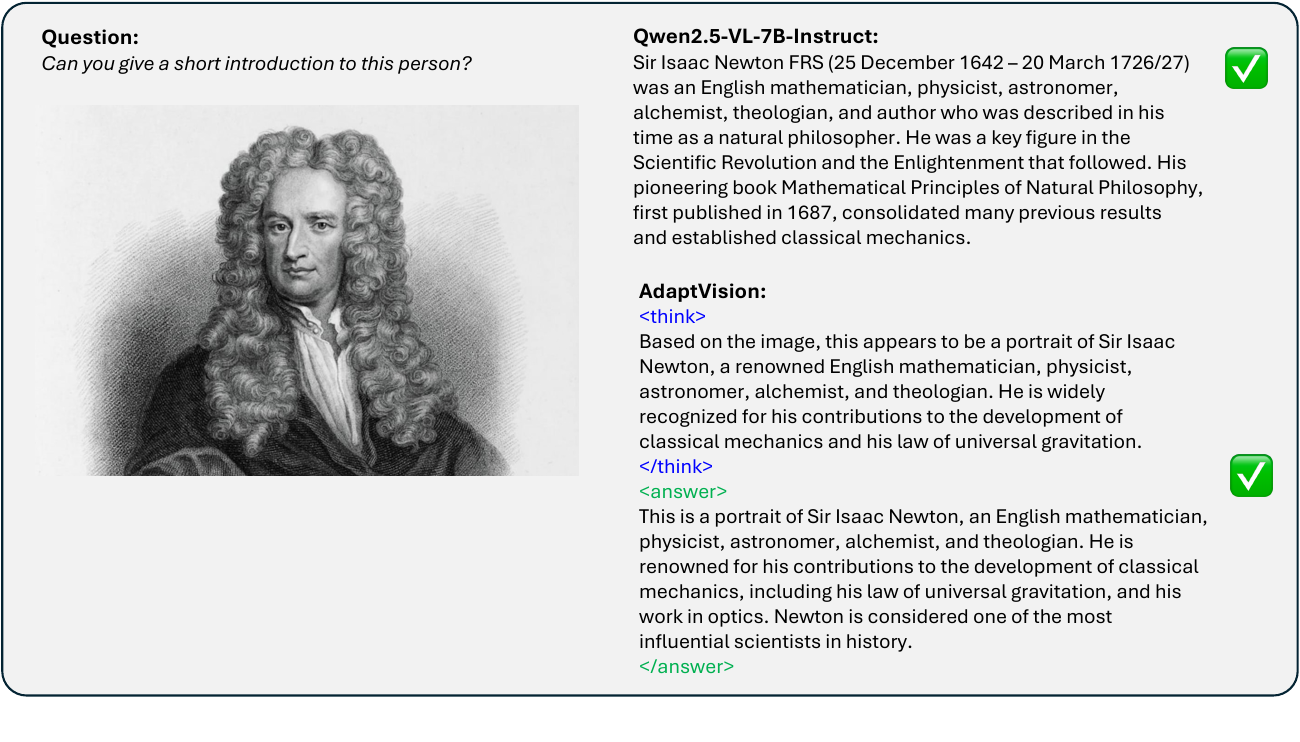}
    \end{center}
    \caption{\textbf{Case of direct answer in AdaptVision.} }
    \label{fig:case_direct_answer}
\end{figure}
\begin{figure}[h]
    \begin{center}
        \includegraphics[width=\linewidth]{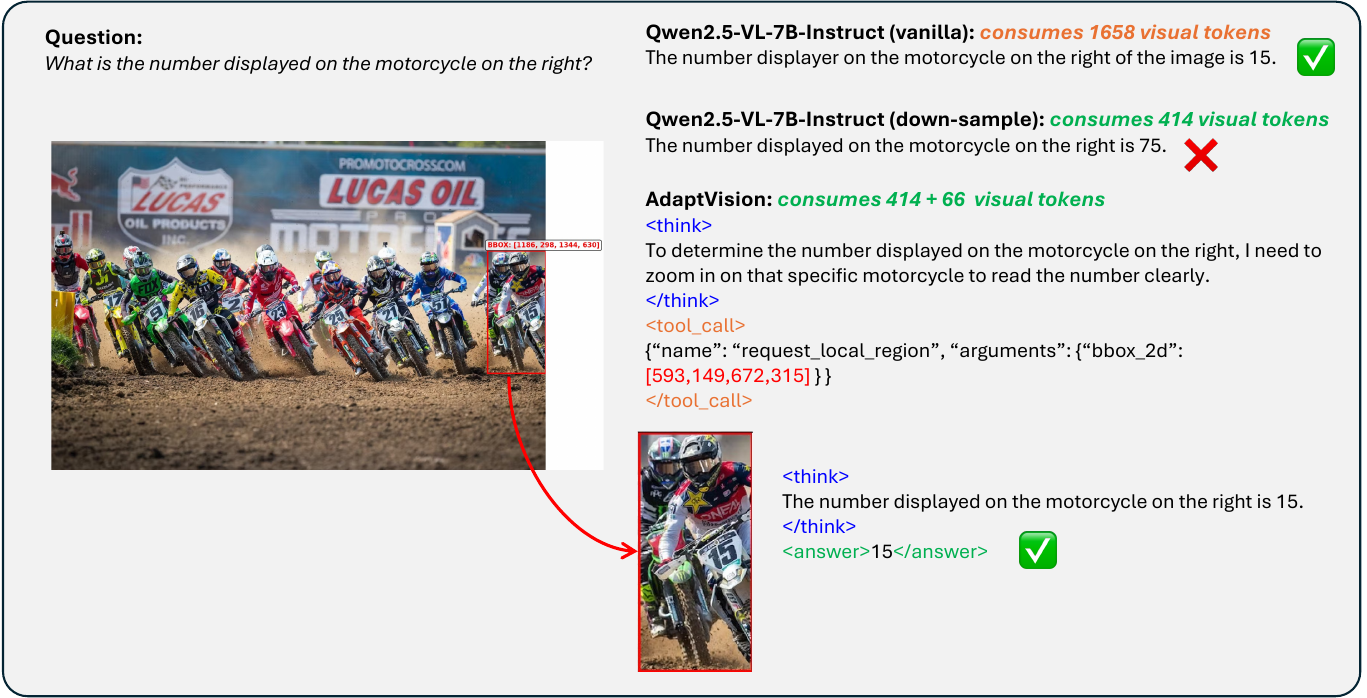}
    \end{center}
    \caption{\textbf{Case of tool call in AdaptVision.} }
    \label{fig:case_tool_call}
\end{figure}

We provide further case studies to illustrate AdaptVision’s adaptive token usage. As shown in Fig.~\ref{fig:case_direct_answer}, in scenarios where a low-resolution image provides enough information, AdaptVision correctly chooses to answer directly—matching the behavior of the Qwen2.5-VL Down-sample model. Conversely, in cases where detailed visual information is essential (Fig.~\ref{fig:case_tool_call}), the Down-sample model often fails due to recognition errors caused by insufficient resolution (e.g., misreading ``15'' as ``75''). Under the same conditions, AdaptVision actively invokes the bounding box tool, accurately localizes informative regions, and produces correct answers with only a marginal increase in visual token consumption relative to the Down-sample model. These examples validate AdaptVision’s ability in coarse-to-fine visual reasoning and its capacity to autonomously tailor visual token usage to each input.

\clearpage


\begin{table}[t]
\centering
\caption{Sensitivity analysis on $\lambda$ and $\alpha$.}
\label{tab:sensitivity}
\resizebox{\linewidth}{!}{
\begin{tabular}{lccccc|cc}
\toprule
 & $\lambda=0.2$ & $\lambda=0.25$ & $\lambda=0.3$ & $\lambda=0.35$ & $\lambda=0.4$ &  $\alpha=1$ & $\alpha=2$ \\
\midrule
RealWorldQA & 64.81 & 66.27 & 67.32 & 66.84 & 66.43 & 66.91 & 67.32 \\
MME & 2368 & 2398 & 2379 & 2378 & 2394 & 2400 & 2379\\
\bottomrule
\end{tabular}
}
\end{table}

\begin{table}[t]
\centering
\caption{Comparison of different reward models.}
\label{tab:performance_comparison}
\resizebox{\linewidth}{!}{
\begin{tabular}{lcccccccc}
\toprule
Model & RealWorldQA & POPE & MME & MathVista & MMVet & \#Token$\downarrow$ \\
\midrule
VisionThink  & 65.6 & 86.3 & 2320 &  62.2 & 61.7 & 51.86\%\\
AdaptVision (GPT-4o)  & 67.32 & 86.8 & 2379 & 65.9 & 64.8 & 30.66\%\\
AdaptVision (Qwen3VL)   & 66.47 & 86.8 & 2313 & 64.7 & 62.5 & 34.18\%\\
\bottomrule
\end{tabular}
}
\end{table}

\section{More Discussion on DTPO}

To help readers to better understand DTPO, we summarize the core contributions of DTPO. First, DTPO decouples advantage estimation by computing distinct advantages for tool and outcome rewards, thereby preventing different rewards from interfering with each other and enabling accurate advantage estimation. By assigning different advantages to distinct tokens, it achieves more precise credit assignment. Second, DTPO decouples policy loss by turns and normalizes the contributions of tool and answer tokens separately, ensuring balanced optimization across tokens with distinct functions. The core contribution of DTPO lies in 1) better credit assignment across different turns and 2) resolving the imbalanced optimization problem across multiple turns. Consequently, DTPO is not limited to the visual reasoning scenario and can be adapted to other multi-turn RL scenarios.


\section{Robustness of DTPO} 

The performance of models trained with different $\lambda$ and $\alpha$ are reported in Table~\ref{tab:sensitivity}. The results show that DTPO is robust to hyperparameters.

\section{Reward Models for AdaptVision}

AdaptVision uses GPT-4o to compute the $R_{crop}$ reward. Here we investigate whether this $R_{crop}$ reward can be computed using a smaller, open-source VLM instead of GPT-4o. 
We conducted an experiment using a smaller open-source VLM, Qwen3-VL-4B-Instruct, to replace GPT-4o as the judge model. From Table~\ref{tab:performance_comparison}, we found that the model still achieves competitive performance and outperforms VisionThink. 

\section{Generalize with Other VLM Architectures}

Our method is architecture-agnostic. It modifies the model's interaction logic and training methodology rather than the model backbone. Thus, it can be applied to other VLM architectures such as Qwen3-VL or InternVL. We chose Qwen2.5-VL-7B-Instruct as our base model because it is a popular and strong open-source VLM baseline at this time. The modular design of AdaptVision ensures its broad applicability, and we encourage future research to extend AdaptVision to alternative VLM architectures.

\section{Extension to Multi-round Tool Calls}

Since our primary goal is to build an efficient VLM, we limit the interaction turns to minimize latency. While we believe that multi-turn tool-use could further improve accuracy by obtaining more fine-grained visual information, it comes at the cost of increased inference time. Nevertheless, we encourage future research to extend AdaptVision to multi-round tool calls to improve accuracy while preserving inference efficiency. For example, applying a deblurring module after zooming in on small objects can significantly improve clarity. This may improve performance in certain cases.

\begin{table*}[t!]
\centering
\caption{\textbf{Performance comparison with previous efficient VLM methods.} Vanilla denotes the Qwen2.5-VL-7B-Instruct model. Down-Sample uses a 1/4-resolution image as input to the Vanilla model. ``\#Token'' indicates the visual token consumption ratio relative to the vanilla model across all benchmarks. ``Avg.'' denotes the average performance relative to the vanilla model on all benchmarks. ``Method (xx\%)'' denotes static methods retaining xx\% visual tokens.}
\resizebox{1\linewidth}{!}{
\begin{tabular}{@{}l|ccccccccc|cc}
\toprule
\multirow{2}{*}{\centering \textbf{Method}} & \textbf{ChartQA} & \textbf{OCRBench} & \textbf{DocVQA} & \textbf{MME} & \textbf{MMVet} & \textbf{RealWorldQA} & \textbf{POPE} & \textbf{MathVista} & \textbf{MathVerse} & \multirow{2}{*}{\centering \textbf{\#Token}$\downarrow$}& \multirow{2}{*}{\centering \textbf{Avg.}$\uparrow$} \\
& test & test & val & test & test & test & test & testmini & testmini & \\
\midrule
\rowcolor{gray!35}
\multicolumn{12}{c}{\textit{Retain 100\% Visual Tokens Across All Benchmarks}}\\
\multirow{2}{*}{\centering Vanilla} & 79.8 & 81.5 & 95.1 & 2316 & 61.6 & 68.6 & 86.7 & 68.2 & 46.3 & \multirow{2}{*}{\centering 100\%} & \multirow{2}{*}{\centering 100\%} \\
& 100\% & 100\% & 100\% & 100\% & 100\% & 100\% & 100\% & 100\% & 100\% \\
\midrule
\rowcolor{gray!35}
\multicolumn{12}{c}{\textit{Retain 25\% Visual Tokens Across All Benchmarks}}\\
\multirow{2}{*}{\centering Down-Sample} & 62.9 & 68.8 & 94.3 & 2270 & 54.5 & 68.8 & 82.8 & 62.2 & 43.1 & \multirow{2}{*}{\centering 25\%} & \multirow{2}{*}{\centering 92.1\%} \\
& 78.8\% & 84.4\% & 99.1\% & 98.0\% & 88.5\% & 100.3\% & 95.5\% & 91.2\% & 93.1\% \\
\midrule
\rowcolor{gray!35}
\multicolumn{12}{c}{\textit{Retain 50\% Visual Tokens Across All Benchmarks}}\\
\multirow{2}{*}{\centering SparseVLM (50\%)} & 73.2 & 75.6 & 66.8 & 2282 & 51.5 & 68.4 & 85.5 & 66.6 & 45.1 & \multirow{2}{*}{\centering 50\%} & \multirow{2}{*}{\centering 92.2\%} \\
& 91.7\% & 92.7\% & 70.2\% & 98.5\% & 83.6\% & 99.7\% & 98.6\% & 97.6\% & 97.4\% \\
\midrule
\multirow{2}{*}{\centering FastV (50\%)} & 72.6 & 75.8 & 93.6 & 2308 & 52.8 & 68.8 & 84.7 & 63.7 & 45.0 & \multirow{2}{*}{\centering 50\%} & \multirow{2}{*}{\centering 95.8\%} \\
& 91.0\% & 93.0\% & 98.4\% & 99.6\% & 85.7\% & 100.3\% & 97.7\% & 93.4\% & 97.2\% \\
\midrule
\multirow{2}{*}{\centering VisionZip (50\%)} & 71.5 & 70.5 & 93.8 & 2209 & 57.0 & 68.6 & 86.3 & 64.1 & 45.1 & \multirow{2}{*}{\centering 50\%} & \multirow{2}{*}{\centering 94.8\%} \\
& 89.6\% & 86.5\% & 98.6\% & 95.4\% & 92.5\% & 100\% & 99.5\% & 93.9\% & 97.4\% \\
\midrule
\rowcolor{gray!35}
\multicolumn{12}{c}{\textit{Retain 70\% Visual Tokens Across All Benchmarks}}\\
\multirow{2}{*}{\centering SparseVLM (70\%)} & 75.8 & 79.3 & 68.7 & 2276 & 53.7 & 68.5 & 85.4 & 66.3 & 45.1 & \multirow{2}{*}{\centering 70\%} & \multirow{2}{*}{\centering 93.6\%} \\
& 94.9\% & 97.3\% & 72.2\% & 98.3\% & 87.2\% & 99.8\% & 98.5\% & 97.2\% & 97.4\% \\
\midrule
\multirow{2}{*}{\centering FastV (70\%)} & 71.2 & 82.2 & 94.4 & 2342 & 56.0 & 68.6 & 85.9 & 65.9 & 46.9 & \multirow{2}{*}{\centering 70\%} & \multirow{2}{*}{\centering 98.4\%} \\
& 96.7\% & 100.8\% & 99.3\% & 101.1\% & 90.9\% & 100\% & 99.1\% & 96.6\% & 101.3\% \\
\midrule
\multirow{2}{*}{\centering VisionZip (70\%)} & 76.8 & 80.9 & 94.5 & 2334 & 60.0 & 68.2 & 86.4 & 68.9 & 45.8 & \multirow{2}{*}{\centering 70\%} & \multirow{2}{*}{\centering 99.1\%} \\
& 96.2\% & 99.3\% & 99.4\% & 100.8\% & 97.4\% & 99.4\% & 99.7\% & 101.0\% & 98.9\% \\
\midrule
\rowcolor{gray!35}
\multicolumn{12}{c}{\textit{Dynamic Methods}}\\
\multirow{2}{*}{\centering VisionThink} & 73.6 & 76.8 & 92.9 & 2320 & 61.7 & 65.6 & 86.3 & 62.2 & 42.5 & \multirow{2}{*}{\centering 52\%} & \multirow{2}{*}{\centering 95.8\%} \\
& 92.2\% & 94.2\% & 97.7\% & 100.2\% & 100.2\% & 95.6\% & 99.5\% & 91.2\% & 91.8\% \\
\midrule
\multirow{2}{*}{\centering VisionThink\textsuperscript{\textdagger}} & 73.88 & 80.8 & 93.7 & 2392 & 60.18 & 68.37 & 86.69 & 65.7 & 45.68 & \multirow{2}{*}{\centering 99\%} & \multirow{2}{*}{\centering 98.4\%} \\
& 92.6\% & 99.1\% & 98.5\% & 103.3\% & 97.7\% & 99.7\% & 100.0\% & 96.3\% & 98.7\% \\
\midrule
\multirow{2}{*}{\centering AdaptVision} & 75.92 & 76.9 & 92.6 & 2379 & 64.8 & 67.32 & 86.8 & 65.9 & 42.3 & \multirow{2}{*}{\centering \textbf{33\%}} & \multirow{2}{*}{\centering \textbf{97.9\%}} \\
& 95.1\% & 94.4\% & 97.4\% & 102.7\% & 105.2\% & 98.1\% & 100.1\% & 96.6\% & 91.4\% \\
\bottomrule
\end{tabular}}
\label{table:more_result}
\end{table*}




\begin{table*}[h]
\centering
\caption{\textbf{Prompt Template for adaptively visual acquisition. } \textcolor{red}{Question} will be replaced with the specific question during training and inference.}
    \begin{tabular}{p{13cm}}
        \toprule
        \textbf{\textit{SYSTEM PROMPT:}} \\
        You are a helpful assistant. \\
        \# Tools \\
        You may call the function tool shown below to assist with the user query. \\
        You are provided with the function signature within \texttt{<tools></tools>} XML tags: \\
        \texttt{<tools>} \\
        \{ \\
        \hspace{10pt} \texttt{"}type\texttt{"}: \texttt{"}function\texttt{"},\\
        \hspace{10pt} \texttt{"}function\texttt{"}:\{ \\
        \hspace{20pt} \texttt{"}name\_for\_human\texttt{"}: \texttt{"}request\_local\_region\texttt{"}, \\
        \hspace{20pt} \texttt{"}name\texttt{"}: \texttt{"}request\_local\_region\texttt{"}, \\
        \hspace{20pt} \texttt{"}description\texttt{"}: \texttt{"}Request a high-resolution local region of the current image and zoom in\texttt{"}, \\
        \hspace{30pt} \texttt{"}parameters\texttt{"}: \{ \\
        \hspace{30pt} \texttt{"}properties\texttt{"}: \{ \\
        \hspace{40pt} \texttt{"}bbox\_2d\texttt{"}: \{ \\
        \hspace{50pt} \texttt{"}type\texttt{"}: \texttt{"}array\texttt{"}, \\
        \hspace{50pt} \texttt{"}items\texttt{"}: \{   \\
        \hspace{60pt} \texttt{"}type\texttt{"}: \texttt{"}integer\texttt{"} \\
        \hspace{50pt} \} \\ 
        \hspace{50pt} \texttt{"}minItems\texttt{"}: 4, \\
        \hspace{50pt} \texttt{"}maxItems\texttt{"}: 4, \\
        \hspace{50pt} \texttt{"}description\texttt{"}:The bounding box of the region to crop, as [x1, y1, x2, y2], where (x1, y1) is the top-left corner of the target region and (x2, y2) is the bottom-right corner of the target region. The bounding box should be in the absolute pixel coordinates of the current image.\texttt{"}, \\
        \hspace{50pt} \} \\ 
        \hspace{40pt} \} \\
        \hspace{30pt} \texttt{"}required\texttt{"}: [\texttt{"}bbox\_2d\texttt{"}], \\
        \hspace{30pt} \texttt{"}type\texttt{"}: \texttt{"}object\texttt{"}, \\
        \hspace{20pt} \}, \\
        \hspace{10pt} \texttt{"}args\_format\texttt{"}: \texttt{"}Format the arguments as a JSON object.\texttt{"} \\
        \hspace{10pt} \} \\
        \} \\
        \texttt{</tools>} \\
        For each function call, return a json object with the function name and the corresponding argument within \texttt{<tool\_call></tool\_call>} XML tags: \\
        \texttt{<tool\_call>} \\
        \{\texttt{"}name\texttt{":<}function-name\texttt{>}, \texttt{"}arguments\texttt{"}:\texttt{<}args-json-object\texttt{>}\} \\
        \texttt{</tool\_call>} \\
        \midrule
        \textbf{\textit{USER PROMPT:}} \\
        Answer the question based on the image provided. You must conduct reasoning within \textcolor{orange}{\texttt{<think>}} and \textcolor{orange}{\texttt{</think>}} first in each of your reasoning steps. You may call ONE function tool per step to help you better solve the problem. Place the function tool within \textcolor{purple}{\texttt{<tool\_call>}} and \textcolor{purple}{\texttt{</tool\_call>}} at the end of each step to perform a function call. You should continue your reasoning process within \textcolor{orange}{\texttt{<think>}} and \textcolor{orange}{\texttt{</think>}} based on the content returned by the function tool. Once you confirm your final answer, place the final answer inside \textcolor{red}{\texttt{<answer>}} and \textcolor{red}{\texttt{</answer>}}.\ For mathematical or multiple-choice problem, wrap the answer value or choice with \textcolor{green}{\texttt{\textbackslash boxed\{\}}}. Here is the image and question: \textcolor{red}{Question}.\\
        \bottomrule
\end{tabular}
\label{tab:vlm_prompt}
\end{table*}




\begin{table*}[h]
    \centering
    \caption{\textbf{Prompt Template for LLM as Final Answer Judge.} \textcolor{red}{Question}, \textcolor{red}{Ground Truth} and \textcolor{red}{Prediction}  are dynamically replaced with the specific question, ground truth and model prediction during evaluation.}
    
    \begin{tabular}{p{13cm}}
        \toprule
        \textbf{\textit{SYSTEM PROMPT:}} \\
        You are an intelligent chatbot designed for evaluating the correctness of generative outputs for question-answer pairs.\\
        Your task is to compare the predicted answer with the correct answer and determine if they match meaningfully. Here's how you can accomplish the task:\\
        INSTRUCTIONS: \\
        - Focus on the meaningful match between the predicted answer and the correct answer.\\
        - Consider synonyms or paraphrases as valid matches.\\
        - Evaluate the correctness of the prediction compared to the answer. \\
        \midrule
        \textbf{\textit{USER PROMPT:}} \\
        I will give you a question related to an image and the following text as inputs:\\
        1. **Question Related to the Image**: \textcolor{red}{Question}\\
        2. **Ground Truth Answer**: \textcolor{red}{Ground Truth}\\
        3. **Model Predicted Answer**: \textcolor{red}{Prediction}\\
        Your task is to evaluate the model's predicted answer against the ground truth answer, based on the context provided by the question related to the image. Consider the following criteria for evaluation:\\
        - **Relevance**: Does the predicted answer directly address the question posed, considering the information provided by the given question?\\
        - **Accuracy**: Compare the predicted answer to the ground truth answer. You need to evaluate from the following two perspectives:\\
        (1) If the ground truth answer is open-ended, consider whether the prediction accurately reflects the information given in the ground truth without introducing factual inaccuracies. If it does, the prediction should be considered correct.\\
        (2) If the ground truth answer is a definitive answer, strictly compare the model's prediction to the actual answer. Pay attention to unit conversions such as length and angle, etc. As long as the results are consistent, the model's prediction should be deemed correct.\\
        **Output Format**:\\
        Your response should include an integer score indicating the correctness of the prediction: 1 for correct and 0 for incorrect. Note that 1 means the model's prediction strictly aligns with the ground truth, while 0 means it does not.\\
        The format should be Score: 0 or 1\\
        \bottomrule
    \end{tabular}
\label{tab:llm_as_answer_judge}
\end{table*}




\begin{table*}[h]
    \centering
    \caption{\textbf{Prompt Template for Judging the Correctness of Bounding Box.} \textcolor{red}{Question} are dynamically replaced with the specific question during evaluation.}
    \begin{tabular}{p{13cm}}
        \toprule
        \textbf{\textit{SYSTEM PROMPT:}} \\
        **Your Role:** You are an AI agent that identifies relevant visual evidence. \\
        **Your Goal:** Determine if an image CROP contains the **primary subject** of a given question. \\
        **Your Golden Rule:** Your main task is to check for **presence**, not completeness. As long as the main object or area the question is asking about is clearly visible in the crop, it is considered relevant. \\
        **Criteria for 'Score: 0' (Strictly Enforced):** \\
        - The core subject of the question is completely absent from the image. \\
        - The image is so blurry or corrupted that the subject is **unrecognizable**. \\
        - The image shows something completely unrelated (e.g. question is about a car, image shows a tree). \\
        **Your Task:** \\
        Now, analyze the user-provided image and question following this exact process. Your response MUST only contain 'Score: 1' or 'Score: 0'. \\
        \midrule
        \textbf{\textit{USER PROMPT:}} \\
        Given a question and a cropped image region, answer with 'Score: 1' if the cropped region provide information to answer the question, otherwise answer 'Score: 0'. Question: \textcolor{red}{Question}." \\
        \bottomrule
    \end{tabular}
\label{tab:judge_bbox_correctness}
\end{table*}




\end{document}